\documentclass[jou,apacite]{apa6}
\newcommand{\jou}{1}

\usepackage{ifthen}
\usepackage{xr}

\ifthenelse{\equal{\jou}{1}}{
}{
	\linespread{1.5}
}

\usepackage{csquotes}
\usepackage{color}

\renewcommand{\subsubsection}[1]{\textbf{#1}\hspace{0.3cm}}

\title{Cognitive science in the era of artificial intelligence: \\A roadmap for reverse-engineering the infant language-learner$^*$}

\authornote{* Author's reprint of Dupoux, E. (2018). Cognitive science in the era of artificial intelligence: A roadmap for reverse-engineering the infant language learner. \textit{Cognition}, \textbf{173}, 43-59, doi: 10.1016/j.cognition.2017.11.008.}

\shorttitle{Reverse-engineering language development}

\author{Emmanuel Dupoux}
\affiliation{EHESS, ENS, PSL Research University, CNRS, INRIA, Paris, France\\ emmanuel.dupoux@gmail.com, www.syntheticlearner.net
	\ifthenelse{\equal{\jou}{1}}{}{~\\
		~\\
		~\\
		~\\
		~\\
		\begin{flushleft}
		\end{flushleft}
	}
}

\leftheader{E. Dupoux}
\begin{document}

	\newcommand{\citeM}[1]{(REF #1)}

	\maketitle    
	
	%%%%%% BEGINCUT

    \ifthenelse{\equal{\jou}{0}}{\newpage}{}

	\section*{Abstract}

Spectacular progress in the information processing sciences (machine learning, wearable sensors) promises to revolutionize the study of cognitive development. Here, we analyse the conditions under which  'reverse engineering' language development, i.e., building an effective system that mimics infant's achievements, can contribute to our scientific understanding of early language development. We argue that, on the computational side, it is important to move from toy problems to the full complexity of the learning situation, and take as input as faithful reconstructions of the sensory signals available to infants as possible. On the data side, accessible but privacy-preserving repositories of home data have to be setup. On the psycholinguistic side, specific tests have to be constructed to benchmark humans and machines at different linguistic levels.  We discuss the feasibility of this approach and present an overview of current results.
	
	\section*{Keywords}
	Artificial intelligence, speech, psycholinguistics, computational modeling, corpus analysis, early language acquisition, infant development, language bootstrapping, machine learning.
	
\section*{Highlights}
\begin{itemize}

\item Key theoretical puzzles of infant language development are still unsolved.
\item A roadmap for reverse engineering infant language learning using AI is proposed.
\item AI algorithms should use realistic input (little or no supervision, raw data).
\item Realistic input should be obtained by large scale recording in ecological environments.
\item Machine/humans comparison should be run on a benchmark of psycholinguistic tests.
\end{itemize}
\ifthenelse{\equal{\jou}{0}}{\newpage}{}

\setcounter{secnumdepth}{2} % in order to number the sections and subsections

\section{Introduction}\label{section:intro}

In recent years, artificial intelligence (AI) has been hitting the headlines with impressive achievements at matching or even beating humans in complex cognitive tasks 
(playing go or video games: \citeNP{mnih_2015,silver_2016}; processing speech and natural language: \citeNP{amodei16,ferrucci_2012}; recognizing objects and faces: \citeNP{he_2015, lu_2014}) and promising a revolution in manufacturing processes and human society at large. These successes 
show that with statistical learning techniques, powerful computers and large amounts of data, it is possible to mimic important components of human cognition. 
Shockingly, some of these achievements have been reached by throwing out some of the classical theories in linguistics and psychology, and by training relatively unstructured neural network systems on large amounts of data. 
What does it tell us about the underlying psychological and/or neural processes that are used by humans to solve these tasks? Can AI provide us with scientific insights about human learning and processing?

Here, we argue that developmental psychology and in particular, the study of language acquisition is one area where, indeed, AI and machine learning advances can be transformational, provided that the involved fields make significant adjustments in their practices in order to adopt what we call the \textit{reverse engineering approach}. Specifically:

\begin{displayquote}
The reverse engineering approach to the study of infant language acquisition consists in constructing \textit{scalable} computational systems that can, when fed with \textit{realistic} input data, \textit{mimic} language acquisition as it is observed in infants. 
\end{displayquote}

The three italicised terms will be discussed at length in subsequent sections of the paper. For now, only an intuitive understanding of these terms will suffice. 
The idea of using machine learning or AI techniques as a means to study child's language learning is actually not new \cite<to name a few:>{kelley_1967,anderson_1975,berwick_85,rumelhart_87,langley_87} although relatively few studies have concentrated on the early phases of language learning \cite<see>[for a pioneering collection of essays]{brent_96}. What is new, however, is that whereas previous AI approaches were limited to proofs of principle on toy or miniature languages, modern AI techniques have scaled up so much that end-to-end language processing systems working with real inputs are now deployed commercially. This paper examines whether and how such unprecedented change in scale could be put to use to address lingering scientific questions in the field of language development.

The structure of the paper is as follows: In Section~\ref{section:puzzles}, we present two deep scientific puzzles that large scale modeling approaches could in principle address: solving the bootstrapping problem, accounting for developmental trajectories. In Section~\ref{section:past}, we review past theoretical and modeling work, showing that these puzzles have not, so far, received an adequate answer. In Section~\ref{section:requirements}, we argue that to answer them with reverse engineering, three requirements have to be addressed: (1) modeling should be computationally scalable, (2) it should be done on realistic data, (3) model performance should be compared with that of humans. In Section \ref{section:deep}, recent progress in AI is reviewed in light of these three requirements. In Section~\ref{section:feasibility}, we assess the feasibility of the reverse engineering approach and lay out the road map that has to be followed to reach its objectives%. %In Section~\ref{section:results} we show that even before these roadblocks are lifted, interesting results can be obtained%. 

, and we conclude in Section~\ref{section:conclusion}.

\section{Two deep puzzles of early language development}\label{section:puzzles}

Language development is a theoretically important subfield within the study of human cognitive development for the following three reasons:

First, the linguistic system is uniquely  \emph{complex}: mastering a language implies mastering a combinatorial sound system (phonetics and phonology), an open ended morphologically structured lexicon, and a compositional syntax and semantics \cite<e.g.,>{jackendoff_97}. No other animal communication system uses such a complex multilayered organization \cite{hauser_2002}. On this basis, it has been claimed that humans have evolved (or acquired through a mutation) an innately specified computational architecture to process language \cite<see>{chomsky_65,steedman_14}. 

Second, the overt manifestations of this system are extremely \emph{variable} across languages and cultures. Language can be expressed through the oral or manual modality. In the oral modality, some languages use only 3 vowels, other more than 20. Consonants inventories vary from 6 to more than 100. Words can be mostly composed of a single syllable (as in Chinese) or long strings of stems and affixes (as in Turkish). Semantic roles can be identified through fixed positions within constituents, or be identified through functional morphemes, etc. \cite<see >[for a typology of language variation]{song_11}. Evidently, infants acquire the relevant variant through learning, not genetic transmission. 

Third, the human language capacity can be viewed as a finite computational system with the ability to generate a (virtual) infinity of utterances. This turns into a \emph{learnability problem} for infants: on the basis of finite evidence, they have to induce the (virtual) infinity corresponding to their language. As has been discussed since Aristotle, such induction problems do not have a generally valid solution. Therefore, language is simultaneously a human-specific biological trait, a highly variable cultural production, and an apparently intractable learning problem.

Despite these complexities, most infants spontaneously learn their native(s) language(s) in a matter of a few years of immersion in a linguistic environment. The more we know about this simple fact, the more puzzling it appears. Specifically, we outline two deep scientific puzzles that a reverse engineering approach could, in principle help to solve: solving the bootstrapping problem and accounting for developmental trajectories. The first puzzle relates to the ultimate outcome of language learning: the so-called \textit{stable state}, defined here as the stabilized language competence in the adult. The second puzzle relates to what we know of the intermediate steps in the acquisition process, and their variations as a function of language input.\footnote{The two puzzles are not independent as they are two facets of the same phenomenon. In practice,  proposals for solving the bootstrapping problem may offer insights about the observed trajectories. Vice-versa, data on developmental trajectories may provide more manageable subgoals for the difficult task of solving the bootstrapping problem.}

\subsection{Solving the bootstrapping problem}
The stable state is the operational knowledge which enables adults to process a virtual infinity of utterances in their native language. The most articulated description of this stable state has been offered by theoretical linguistics; it is viewed as a grammar comprising several components: phonetics, phonology, morphology, syntax, semantics, pragmatics. 

The \textit{bootstrapping problem} arises from the fact these different components appear \emph{interdependent} from a learning point of view. For instance, the phoneme inventory of a language is defined through pairs of words that differ minimally in sounds (e.g., "light" vs "right"). This would suggest that to learn phonemes, infants need to first learn words. However, from a processing viewpoint, words are recognized through their phonological constituents \cite<e.g.,>{cutler_12}, suggesting that infants should learn phonemes before words. Similar paradoxical co-dependency issues have been noted between other linguistic levels (for instance, syntax and semantics: \citeNP{pinker_87}, prosody and syntax: \citeNP{morgan_96}). %, see Figure \ref{fig:tableau}b). 
In other words, in order to learn any one component of the language competence, many others need to be learned first, creating apparent circularities. 

The bootstrapping problem is further compounded by the fact that infants do not have to be taught formal linguistics or language courses to learn their native language(s). As in other cases of animal communication, infants \emph{spontaneously} acquire the language(s) of their community by merely being immersed in that community \cite{pinker_language_1994}. Experimental and observational studies have revealed that infants start acquiring elements of their language (phonetics, phonology, lexicon, syntax and semantics) even before they can talk \cite{jusczyk_1997,hollich_2000,werker_2005}, and therefore before parents can give them much feedback about their progress into language learning. This suggests that language learning (at least the initial bootstrapping steps) occurs largely \emph{without supervisory feedback}.\footnote{Even in later acquisitions, the nature, universality and effectiveness of corrective feedback of children's outputs has been debated \cite<see>{brown_73,pinker_89, marcus_93, chouinard_03,saxton_97,clark_lappin_11}.} %\footnote{Besides uncertainties on the cross-cultural prevalence of corrective feedback is the fact that during the first few years of life, infant's language productions lag significantly behind their comprehension; it is therefore not clear that even if feedback is prevalent and effective, it is truly informative as regards the learnability of the underlying linguistic structures.}. 

The reverse engineering approach has the potential of solving this puzzle by providing a computational system that can demonstrably bootstrap into language when fed with similar, supervisory poor, inputs\footnote{A sucessful system may not necessarily have the same architecture of components as described by theoretical linguists. It just needs to behave as humans do, i.e., pass the same behavioral tests. More on this in section \ref{section:eval}.}.

\begin{figure*}[t]
	\centering
	\includegraphics[height=8cm]{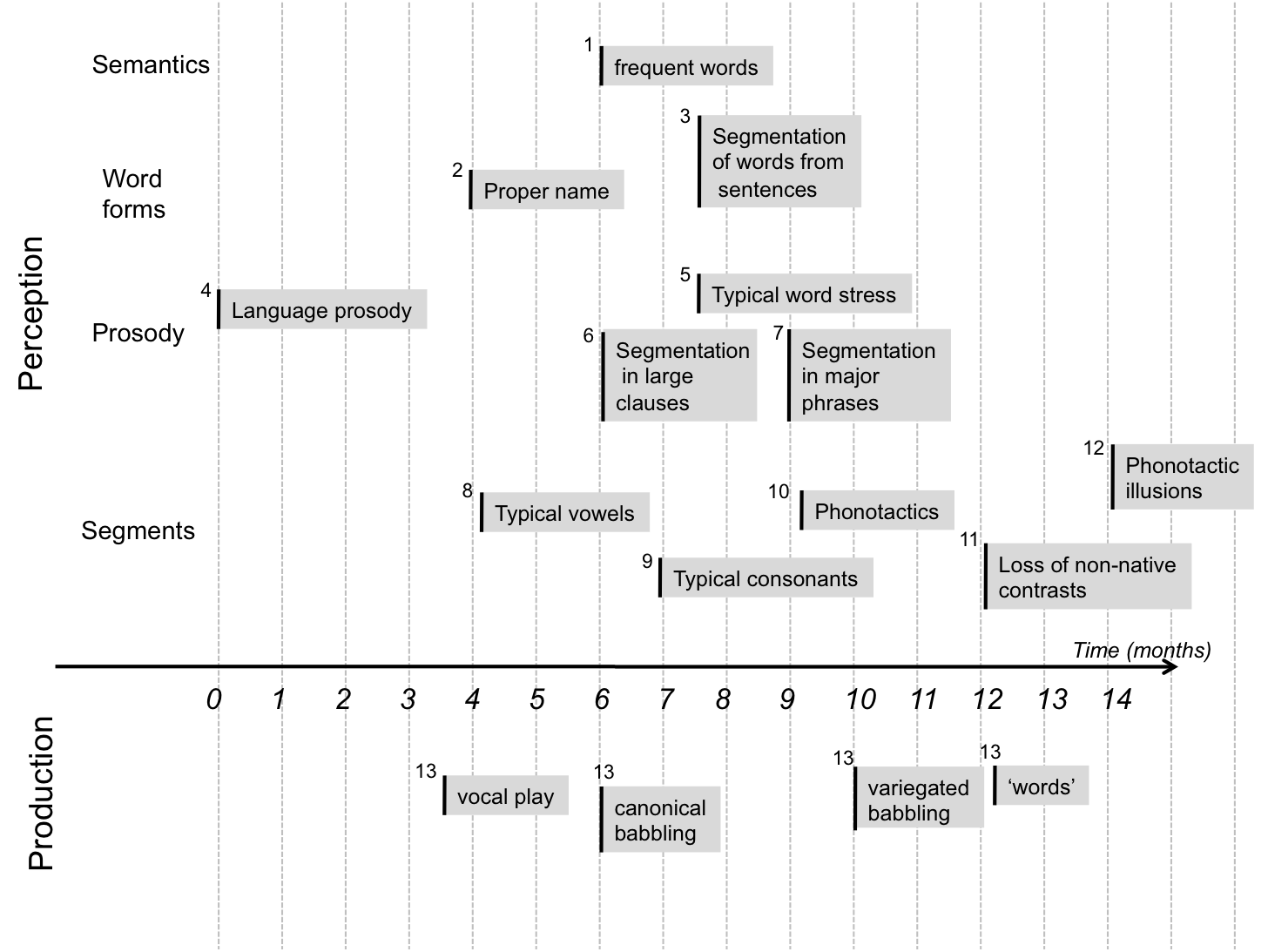}
	
	%	a{\label{fig1:a}\includegraphics[height=8cm]{timeline2.png}}\hfill
	%	b{\label{fig1:b}\includegraphics[height=6 cm]{dependencies.png}}
	\caption{Sample studies illustrating the \textit{time line }of infant's language development. The left edge of each box is aligned to the earliest age at which the result has been documented. $^{1}$ Tincoff \& Jusczyk, (1999); Bergelson \& Swingley, (2012); $^{2}$ Mandel et al. (1995); $^{3}$ Jusczyk \& Aslin (1995) $^{4}$ Mehler et al. (1988) $^{5}$ Jusczyk et al. (1999) $^{6}$ Hirsh-Pasek et al. (1987) $^{7}$  Jusczyk et al (1992) $^{8}$ Kuhl et al. (1992) $^{9}$ Eilers et al. (1979) $^{10}$ Jusczyk et al. (1993) $^{11}$ Werker \& Tees (1984) $^{12}$ Mazuka et al. (2011) $^{13}$ Stark (1980).}
	
	\label{fig:tableau}
\end{figure*}
\nocite{tincoff_1999,bergelson_12,mandel1995,jusczyk_1995}
\nocite{mehler_1988,jusczyk_1999,hirsh_1987,jusczyk_1992,kuhl_1992}
\nocite{eilers_1979,jusczyk_93,werker_84,mazuka_2011,stark_1980}

\subsection{Accounting for developmental trajectories}

In the last forty years, a large body of empirical work has been collected regarding infant's language achievements during their first years of life. This work has only added more puzzlement.

First, given the multi-layered structure of language, one could expect a stage-like developmental tableau where acquisition would proceed as a discrete succession of learning phases organized logically or hierarchically (e.g., building linguistic structure from the low level to the high levels). This is not what is observed  (see Figure \ref{fig:tableau}). For instance, infants start differentiating native from foreign consonants and vowels at 6 months, but continue to fine tune their phonetic categories well after the first year of life \cite<e.g., >{sundara_2006}. However, they start learning about the sequential structure of phonemes (phonotactics, see \citeNP{jusczyk_93}) way \textit{before} they are done acquiring the phoneme inventory \cite{werker_84}. Even before that, they start acquiring the meaning of a small set of common words \cite<e.g.>{bergelson_12}. In other words, instead of a stage-like developmental tableau, the evidence shows that acquisition takes places at all levels more or less simultaneously, in a \emph{gradual} and largely \emph{overlapping} fashion.

Second, observational studies have revealed considerable \emph{variations} in the \emph{amount of language input} to infants across cultures \cite{shneidman_12} and across socio-economic strata \cite{hart_95}, some of which can exceed an order of magnitude (\citeNP{weisleder_13}, p. 2146; \citeNP{cristia:2017}; see also Supplementary Section S1). These variations do impact language achievement as measured by vocabulary size and syntactic complexity \cite[among others]{hoff_03,huttenlocher_10,pan_05,rowe_09}, but at least for some markers of language achievement, the differences in outcome are much less extreme than the variations in input. For canonical babbling, for instance, an order of magnitude would mean that some children start to babble at 6 months, and others at 5 years!  The observed range is between 6 and 10 months, less than a 1 to 2 ratio. Similarly, reduced range of variations are found for the onset of word production and the onset of word combinations. This suggests a surprising level of \emph{resilience} in language learning, i.e., some minimal amount of input is sufficient to trigger certain landmarks.

The reverse engineering approach has the potential of accounting for this otherwise perplexing developmental tableau, and provide quantitative predictions both across linguistic levels (gradual overlapping pattern), and cultural or individual variations in input (resilience).

\section{Standard approaches to language development}\label{section:past}

It is impossible in limited space to do justice to the rich and diverse sets of viewpoints that have been proposed to account for language development. Instead, the next sections will present a non exhaustive selection of four research strands which draw their source of inspiration from a mixture of psycholinguistics, formal linguistics and computer science, and which share some of the explanatory goals of the reverse engineering approach. The argument will be that even though these strands provide important insights into the acquisition process, they still fall short of accounting for the two puzzles presented in Section~\ref{section:puzzles}.

\subsection{Psycholinguistics: Conceptual frameworks}

Within developmental psycholinguistics, \emph{conceptual frameworks} have been proposed to account for key aspects of the bootstrapping problem and developmental trajectories (see Table \ref{tab:conceptualf} for a non exhaustive sample). 

Specifically adressing the bootstrapping problem, some frameworks build on systematic correlations between linguistic levels,  e.g., between syntactic and semantic categories (syntactic bootstrapping: \citeNP{gleitman_1990}; semantic bootstrapping: \citeNP{grimshaw_81,pinker_1984}), or between prosodic boundaries and syntactic ones (prosodic bootstrapping: \citeNP{morgan_96,christophe_08}. %,christophe:1997}).),  which can help breaking logical circularities: , and  
Others endorse Chomsky's \citeyear{chomsky_65} hypothesis that infants are equipped with an innate Language Acquisition Device %(LAD)
which constrains the hypothesis space of the learner, enabling acquisition in the presence of scarse or ambiguous input \cite{crain_91,lidz_2015}.

Other conceptual frameworks focus on key aspects of developmental trajectories (patterns across ages, across languages, across individuals), offering overarching architectures or scenarios that integrate many empirical results. Among others, the competition model: \citeNP{bates_1987,macwhinney_87} ; WRAPSA: \citeNP{jusczyk_1997};  the emergentist coalition model: \citeNP{hollich_2000}; PRIMIR: \citeNP{werker_2005}; usage-based theory: \citeNP{tomasello_03}. Each of these frameworks propose a collection of mechanisms linked to the linguistic input and/or the social environment of the infant to account for developmental trajectories.

\begin{table*}[]
		\small
	\centering
	\caption{Non-exhaustive sample of conceptual frameworks accounting for aspects of early language acquisition. (BP: Bootstrapping Problem; DT: Developmental Trajectories)}
	\label{tab:conceptualf}

\begin{tabular}{p{0.18\linewidth}p{0.2\linewidth} p{0.24\linewidth} p{0.26\linewidth}}
	\hline 
	Conceptual framework & Reference & Puzzle addressed  (age range)& Proposed mechanism\\ 
	\hline
	Semantic Bootstrapping & \citeNP{pinker_1984} &  BP: syntax  (production, 18 mo-4 y)& inductive biases based on syntax/semantic correlations\\ 
	Syntactic Bootstrapping & \citeNP{gleitman_1990} &  BP: verb semantics  (perception, 18 mo-4 y)& inductive biases on syntax/semantic correlations\\ 
	Prosodic  Bootstrapping & \citeNP{morgan_96} &  BP: word segmentation, part of speech, syntax (perception, 9-18 mo)& inductive biases on prosodic/lexicon/syntax correlations\\ 
	Knowledge-driven LAD&\citeNP{lidz_2015}	& BP: syntax (perception \& production, 18 mo -8 y) &  perceptual intake; Universal Grammar; inference engine\\  
	WRAPSA & \citeNP{jusczyk_1997}   &  DT:  phonetic categories, word segmentation (perception, 0-12 mo) & auditory processing; syllable segmentation; attentional weighting; pattern extraction; exemplar theory\\  
	PRIMIR   & \citeNP{werker_2005}     & DT: phonetic, speaker, phonological and semantic categories (perception, 0-2 y) & examplar theory, statistical clustering, associative learning, attentional dynamic filters\\ 
	Competition Model & \citeNP{bates_1987,macwhinney_87}  &  DT: syntax (production, 18 mo - 4 y)& competitive learning (avoidance of synonymy)\\ 
   Usage-Based Theory & \citeNP{tomasello_03} &  DT: semantics,  syntax (perception \& production, 9 mo - 6 y)& construction grammar; intention reading; analogy; competitive learning; distributional analysis\\ 
   \hline 
\end{tabular} 
\end{table*}

While these conceptual framework are very useful in summarizing and organizing a vast amount of empirical results, and offer penetrating insights, they are not specific enough to address our two scientific puzzles. They tend to refer to mechanisms using verbal descriptions (\emph{statistical learning}, \emph{rule learning}, \emph{abstraction}, \emph{grammaticalization}, \emph{analogy}) or boxes and arrows diagrams.  This type of presentation may be intuitive, but also vague. The same description may correspond to many different computational mechanisms which would yield different predictions. These frameworks are therefore difficult to distinguish from one another empirically, or for the most descriptive ones, impossible to disprove.  In addition, because they are not formal, one cannot demonstrate that these models can effectively solve the language bootstrapping problem. Nor do they provide quantitative predictions about the observed resilience in developmental trajectories or their variations as a function of language input at the individual, linguistic or cultural level.

\subsection{Psycholinguistics: Artificial language learning}

Psycholinguists sometimes supplement conceptual frameworks with propositions for specific \emph{learning mechanisms} which are tested using an artificial language paradigm. As an example, a mechanism based on the tracking of statistical modes in phonetic space has been proposed to underpin phonetic category learning in infancy. It was tested in infants through the presentation of a simplified language (a continuum of syllables between /da/ and /ta/) where the statistical distribution of acoustic tokens was controlled \cite{maye_2002}. It was also modeled computationally using unsupervised clustering algorithms and tested using simplified corpora or synthetic data \cite{vallabha_2007,mcmurray_2009}. A similar double-pronged approach (experimental and modeling evidence) has been conducted for other mechanisms: word segmentation based on transition probability  \cite{saffran_1996,daland_2011},  word meaning learning based on cross situational statistics \cite{yu_2007,smith_2011,siskind_1996}, semantic role learning based on syntactic cues \cite{connor_2013}, etc.

Although studies with artificial languages are useful to discover candidate learning algorithms which could be incorporated in a global architecture, the algorithms proposed have only been tested on toy or artificial languages; there is therefore no guarantee that they would actually work when faced with realistic corpora that are both very large and very noisy (see \ref{section:theorybenefit}).

\subsection{Formal linguistics: learnability studies}
Even though much of current theoretical linguistics is devoted to the study of the language competence in the stable state, very interesting work has also been conducted in the area of formal models of \emph{grammar induction}. These models propose algorithms that are provably powerful enough to learn a fragment of grammar given certain assumptions about the input.  For instance, \citeA{tesar_1998} proposed an algorithm that provided pairs of surface and underlying word forms can learn the phonological grammar (see also \citeNP{magri_2015}). Similar learnability assumptions and results have been obtained for stress systems \cite{dresher_1990,tesar_2000}. For learnability results of syntax, see \citeA{clark_lappin_11}.

These models establish important learnability results, and in particular, demonstrate that under certain hypotheses, a particular class of grammar is learnable.  What they do not demonstrate however is that these hypotheses are met for infants. In particular, most grammar induction studies assume that infants have an error-free, adult-like symbolic representation of linguistic entities (e.g., phonemes, phonological features, grammatical categories, etc). Yet, perception is certainly not error-free, and it is not clear that infants have adult-like symbols, and if they do, how they acquired them.

In other words, even though these models are more advanced than psycholinguistic models in formally addressing the effectiveness of the proposed learning algorithms, it is not clear that they are solving the same bootstrapping problem than the one faced by infants. In addition, they typically lack a connection with empirical data on developmental trajectories.\footnote{A particular difficulty of formal models which lack a processing component is to account for the observed discrepancies between the developmental trajectories in perception (e.g. early phonotactic learning in 8-month-olds) and production (slow phonotactic learning in one to 3-year-olds).}

\subsection{Developmental artificial intelligence}

The idea of using computational models to shed light on language acquisition is as old as the field of cognitive science itself, and a complete review would be beyond the scope of this paper. We mention some of the landmarks in this field which we refer to as \textit{developmental AI}, separating three learning subproblems: syntax, lexicon, and speech. 

Computational models of syntax learning in infants can be roughly classified into two strands, one that learns  from strings of words alone, and one that additionally uses a conceptual representation of the utterance meaning. The first strand is illustrated by \citeA{kelley_1967}. It views grammar induction as a problem of representing the input corpus with a grammar in the most compact fashion, using both a priori constraints on the shape and complexity of the grammars and a measure of fitness of the grammar to the data (see \citeNP{de_marcken_1996} for a probabilistic view). The first systems used artificial input (generated by a context free grammar) and part-of-speech tags (nouns, verbs, etc.) were provided as side-information.  Since then, manual tagging has been replaced by automatic tagging using a variety of approaches (see \citeNP{christodoulopoulos_2010} for a review), and artificial datasets have been replaced by naturalistic ones (see \citeNP{dulizia_2011}, for a review).  The second strand can be traced back to \citeA{siklossy_1968}, and makes the radically different hypothesis that language learning is essentially a translation problem: children are provided with a parallel corpus of speech in an unknown language, and a conceptual representation of the corresponding meaning. The Language Acquisition System (LAS) of  \citeA{anderson_1975} is a good illustration of this approach. It learns context-free parsers when provided with pairs of representations of meaning (viewed as logical form trees) and sentences (viewed as a string of words, whose meaning are known). Since then, algorithms have been proposed to learn directly the meaning of words (e.g., cross-situational learning, see \citeNP{siskind_1996}), context-free grammars have been replaced by more powerful ones (e.g. probabilistic Combinatorial Categorical Grammar), and sentence meaning has been replaced by sets of candidate meanings with noise (although still generated from linguistic annotations) \cite<e.g.,>{kwiatkowski_2012}. Note that both types of models take textual input, and therefore make the (incorrect) assumption that infants are able to represent their input in terms of an error-free segmented string of words. 
  
Computational models of word discovery tackle the problem of segmenting a continuous stream of phonemes into word-like units. One idea is to use distributional properties that distinguish within word and between word phoneme sequences \cite{harris_1954,elman_1990,christiansen_2005}. A second idea is to simultaneously build a lexicon and segment sentences into words \cite{olivier_1968,de_marcken_1996,goldwater_2007}. These ideas are now frequently combined \cite{brent_1996,johnson_2008}.  In addition, segmentation models have been augmented by jointly learning the lexicon and morphological decomposition \cite{johnson_2008,botha_2013}, or tackling phonological variation through the use of a noisy channel model \cite{elsner_2012}. Note that all of these studies assume that speech is represented as an error-free string of adult-like phonemes, an assumption which cannot apply to early language learners. 

Finally, a few computational model have started to address language learning from raw speech. These have either concerned the discovery of phoneme-sized units, the discovery of words, or both. Several ideas have been proposed to discover phonemes from the speech signal (self organizing maps: \citeNP{kohonen_1988}; clustering: \citeNP{pons_2013}; auto-encoders: \citeNP{badino_2014}; HMMs: \citeNP{siu_2013}; etc.). Regarding words, \citeA{roy_2002} proposed a model that learn both to segment continuous speech into words and map them to visual categories (through cross situational learning). This was one of the first models to work from a real speech corpus (parents interacting with their infants in a semi-directed fashion), although the model used the output of a supervised phoneme recognizer.  The ACORNS project \cite{boves_2007} used raw speech as input to discover candidate words (\citeNP{ten_bosch_2007}, see also \citeNP{park_2008,muscariello_2009}, etc.), or to learn word-meaning associations (see a review in \citeNP{rasanen_2012}, and a comprehensive model in \citeNP{rasanen2015}), although the speech was collected in the laboratory, not in real life situations.

In sum, developmental AI represents the clearest attempt so far of addressing the full bootstrapping problem. Yet, although one can see a clear progression, from simple models and toy examples, towards more integrative algorithms and more realistic datasets, there is 
still a large gap between models that learn from speech, which are limited to the discovery or phonemes and word forms, and models that learn syntax and semantics, which only work from textual input. Until this gap is closed, it is not clear how the bootstrapping problem as faced by infants can be solved. The research itself is unfortunately scattered in disjoint segments of the literature, with little sharing in algorithms, evaluation methods and corpora, making it difficult to compare the merits of the different ideas and register progress. Finally, even though most of these studies mention  infants as a source of inspiration of the models, they seldom attempt to account for developmental trajectories.

\subsection{Summing up}

\begin{table*}[]
	\centering
	\captionsetup{justification=centering}
	\caption{Summary of different theoretical approaches to the study of early language acquisition.}
	\label{tab:pastwork}
	
	\begin{tabular}{llll}
		\hline 
		       & Effective &Realistic  & {Human/Model}    \\ 	
    	       & Model     & Data        & Comparison           \\

		\hline 
		Conceptual Frameworks      &No (verbal)&  Yes           & No (verbal) \\ 
		Artficial Language Learning &Yes (but not scalable)            &  No            &   Yes          \\ 
		Formal Linguistics               &Existence proof            & Idealized   & In the limit        \\ 
		Developmental AI                &Yes            &  Simplified & Qualitative /     \\ 
		                                            &                 &                  & In the limit      \\ 

		Reverse Engineering            &Yes            & Yes             & Yes               \\ 
		\hline 
	\end{tabular} 
\end{table*}

Psycholinguistic conceptual frameworks capture important insights about language development but are not specified enough to demonstrably solve the bootstrapping problem nor can they make quantitative predictions. Artificial language experiments yield interesting learning mechanisms aimed at explaining experimental data but not necessarily to scale up to larger or more noisy data. These limitations call for the need to develop \textit{effective computational models} that work at scale. Both linguistic models and developmental AI attempt to effectively address the bootstrapping problem, but make unrealistic assumptions with respect to the input data (linguistic models take only symbolic input data, and most developmental AI models take either symbolic data or simplified inputs). As a result, these models may address a different bootstrapping problem than the one faced by infants. This would call for the need to use \textit{realistic data} as input for models. Both linguistic models and developmental AI models take as their gold standard description of the stable state in adults. This may be fine when the objective is to explain ultimate attainment (the bootstrapping problem), but does not enable to connect with learning trajectory data. This would call for a direct \textit{human-machine comparison}, at all ages. %Finally, a problem with much past computational modeling research in general is that even though they enabled quantitative predictions, the resources used were both specific to each study and not distributed freely, making it difficult to compare the different ideas and build on them. This calls for \textit{open sourcing} these resources.

Obiously, the four reviewed research traditions have limits but also address part of the language development puzzles (Table \ref{tab:pastwork}). Before examining how the reverse engineering approach could combine the best of these traditions, we examine next with more scrutiny the requirements they have to meet in order to fully address these puzzles.

\section{The three requirements of the reverse engineering approach}\label{section:requirements}

Here, we argue that to be of scientific import, models of development should (1) go beyond conceptual and box-and-arrow frameworks and be turned into effective, scalable computational systems, (2) go beyond toy data and be fed with realistic input, and (3) be evaluated through human/machine comparisons.

\subsection{Why scalable computational models?}\label{section:model}

Scalable computational systems can provide a proof of principle that the bootstrappping problem can be solved, and generate quantitative predictions.  But there is an even more compelling reason to strive for them:  verbal resoning and toy models tend to badly misjudge how a combination of contradictory tendencies will play out in practice, resulting in sometimes spectacularly incorrect predictions. We illustrate this with three examples.

\subsubsection{'Easy' problems proving difficult.}\label{section:theorybenefit}

How do infant learn phonemes? A popular hypothesis ('\textit{distributional learning}') states that they track the statistical modes of speech sounds to construct phonetic categories \cite{maye_2002}. How do we turn such a verbal description into a scalable algorithm?  

\citeA{vallabha_2007} and \citeA{mcmurray_2009}, among others, have proposed that it can be done with \textit{unsupervised clustering} algorithms.  As it turns out, these algorithms were only validated only on toy data (points in formant space generated from a Gaussian distribution) or on manually obtained measurments. This is a problem because many if not most clustering algorithms are sensitive to data size, variability and dimensionality \cite{fahad_14}. When tested on continuous audio representations which are large, variable and of high dimension, very different result ensue. For instance, \citeA{varadarajan_2008} have shown that a clustering algorithm based on Hidden Markov Models and Gaussian mixtures does not converge on phonetic segments, but rather, on much shorter (30 ms), highly context-sensitive acoustic clusters (see also \citeNP{antetomaso_2017}). This is not surprising given that phonemes are not realized as discrete acoustic events but as complicated overlapping gestures. For instance, a stop consonant surfaces as a burst, a closure, and formant transitions into the next segment.%\footnote{Even though simple clustering models do not learn phonemes when run on speech signals, they may be able to capture some aspects of phonetic learning in infants (see \cite{schatz_17}).}.

This shows that contrary to the  distributional learning hypothesis, finding phonetic units is not only a problem of \textit{clustering}, it is also includes \textit{continuous speech segmentation} and \textit{contextual modeling}. These problems are not independent and have therefore to be addressed jointly by the learning algorithms.  
Despite the optimistic conclusions of \citeA{vallabha_2007} and \citeA{mcmurray_2009}, the unsupervised discovery of phonetic categories is still an unsolved problem in speech technology (see \citeNP{versteegh_2016,dunbar2017}). %Once solve, the resulting algorithms would of course become plausible candidates that instanciate the original hypothesis.

\subsubsection{'Impossible' approaches turning out feasible.}\label{section:theorybenefit}
The second example relates to the popular hypothesis that acquiring the meaning of words is essentially a problem of associating word form to referents in the outside world (or to conceptual representations of these referents; see \citeNP{bloom_2000} for possible learning mechanisms). 

Under such a view, it would seem impossible to learn any word meaning from language input only. 
However, research in natural language processing has shown that it is in fact possible to derive an approximate representation of the word meanings using only coocurrence patterns within the verbal material itself. These distributional techniques \cite{landauer_1997,mikolov_2013} construct vector representation of word meanings  which correlate surprisingly well with human semantic similarity judgments \cite{turney_2010,baroni_2014}\footnote{Interestingly, text-based distributional semantic tend to predict human semantic similarity judgments better than image-based representations \cite{bruni_2012,kiela_2014,silberer_2016}.}.  \citeA{fourtassi2014c}  found %on large corpora of English and Japanese
that it is possible to derive such vectors even without any properly segmented lexicon, and even without adult-like phonetic categories. It turns out that the approximate meaning representation so derived can provide top-down feedback helping clustering phonetic information into phonemes. Thus, computational systems can suggest a priori implausible, but potentially effective, mechanisms. The empirical validity of such mechanisms in infants remains to be tested.%This suggests an interesting complementary mechanism to distributional learning.

\subsubsection{Statistically significant effects ending up unimportant.}\label{section:theorybenefit}
A third example relates to the so-called '\textit{hyperspeech hypothesis}'. It has been proposed that parents adapt their pattern of speech to infants in order to facilitate  perception \cite{fernald_2000}.  \citeA{kuhl_1997} observed that parents tend to increase the separation between point vowels in child directed speech,  possibly making them easier to learn. Yet, \citeA{ludusan_2015} ran a word discovery algorithm borrowed from developmental AI on raw speech and failed to find any difference in word learning between child and adult directed speech; if anything, the former was slightly more difficult. This paradoxical result can be explained by the fact that in child directed speech, parents increase phonetic variability even more than they increase the separation between point vowels, the two effects not only cancel each other out, but even result in a small net \textit{degradation} in category discriminability (\citeNP{martin_2015}; see also \citeNP{mcmurray_2013,guevara_17}). The lesson is that it is only through a completely explicit model that the quantitative effect of linguistic and phonetic variables on learning can be assessed.

\subsection{Why using realistic data?}\label{section:data} 
We turn here to the most controversial of the three requirements: the idea that one should address language learning in its full complexity by running computational models on inputs that are as close as infants' sensory signals as possible.

This may seem an exageration. Simplification is the hallmark of the scientific method, which  usually proceeds by breaking down complicated problems into smaller, more manageable ones. Here, we claim that an exception has to be made for language learnability. Why? In a nutshell: learning is a process whose outcome is exquisitely sensitive to details of the input signal. If one makes even slightly incorrect assumptions about the input of the learning process, one ends up studying a different learning problem altogether. We illustrate this with three cases where simplifications is a learnability game changer.  We conclude that since the learnability-relevant properties of infant's input are currently unknown,  the only possibility left is to go with the real thing.

\subsubsection{Data selection matters.}
The entire set of sensory stimulations available to the child is called the input. The subset of this input which is used to learn about the target language(s) is called the intake. The difference between input and intake defines a \textit{data selection problem} which, we claim, is an important part of the learning problem itself. Unfortunately, many computational models of language acquisition short-circuit the selection problem and use human experts to prepare pre-selected and pre-cleaned data. We illustrate this with three data selection problems.%; in other words, part of the problem is being solved for the algorithm through a preliminary phase of data preparation. 

The first problem relates to defining what counts as linguistic versus non-linguistic information. There is no language-universal answer to this question. For instance, gestures are typically para- or extra-linguistic in communities using oral communication \cite{fowler_91,goldin_05},
but they are the main vehicle for language in sign language \cite{poizner_87}
which is learned by children in deaf or mixed hearing/deaf communities \cite{vancleve_04}.
Within the auditory modality, some vocal sounds like clicks are considered as non-linguistic in many languages, but in others they are used phonologically \cite{best_88}; similarly for phonatory characteristics of vowels like breathiness and creakiness \cite{silverman_95,podesva_2007}. 

The second problem is that even if linguistic and non-linguistic signals are defined for a language, the actual un-mixing of these signals may be difficult. For instance, infants hear a superposition of many audio sources, only some of which contain linguistic signals. Auditory source separation is a computationally difficult problem (untractable in general). In human adults, it is influenced by top-down word recognition (e.g. \citeNP{warren_1970}). In pre-verbal infants such sources of top-down information have themselves to be learned. 

The third problem is that even if non-linguistic signals are separated from linguistic ones, what to do with non-linguistic signals? In most instances, they should be considered as noise and discarded. In other cases, however, they can be useful for language learning. For instance, non-linguistic contextually relevant information in the form of visually perceived objects or scenes may help lexical learning \cite{roy_2002}  or bootstrap syntactic learning (the semantic bootstrapping hypothesis, see \citeNP{pinker_1984}). 
Social signals (eye gaze, touch, etc), have also been taken as crucial for language learning \cite[among others]{tomasello_03,werker_2005}. Here again, the proper channeling of these non-linguistic cues is part of the learning problem.

In brief, data selection is a critical component of a learning problem. It should \textit{not} be performed by the modeler, who has inside information about the target language and culture, but by the model, whose task is precisely to discover it.

\subsubsection{Variability and ambiguity matter.}
Assuming data selection is solved, ambiguity and variability are prevalent properties of language, at all level of structure, from phonetics up to semantics and pragmatics. Yet, many modeling approaches simplify this complexity by replacing real input with synthetic or idealized data. Although doing so is a useful practice to debug algorithms or prove mathematical results, generalizing from the simplified to real input is risky business.

We already discussed how clustering algorithms that discover phonetic categories when run on  synthetic or simplified phonetic data yield much totally different results when run on speech signals. One level up, word segmentation algorithms that recover word boundaries when fed with (errorless) phoneme transcriptions \cite{goldwater_2007} utterly fail when run on speech signals \cite{jansen_2013b,ludusan_14}.  The problem is pervasive. Learning algorithms work because they incorporate models of the shape of the data to be learned. Mismatches between the models and the data will likely result in a learning failure. 

Vice versa, however, oversimplifying the input can make the learning problem harder than it is in reality. As an example, syntax learning models often operate from abstract transcriptions, and as a result ignore prosodic information which could prove useful for the purpose of syntactic analysis, or lexical acquisition \cite<e.g.>{morgan_96,christophe_08,shukla_2011}.

\subsubsection{'Presentation' matters.}
The notion of 'presentation' comes from formal learning theory \cite{gold_1967}. It corresponds to the particular way or order in which a parent  selects his or her language inputs to the child.  There are well known examples where presentation has extreme consequences on what can be learned or not. For instance, if there are no constraints on the order in which environment presents grammatical sentences, then even simple classes of grammars  \cite<e.g., finite state or context free grammars,>{gold_1967} are unlearnable. In contrast, if the environment presents sentences according to a computable process (an apparently innocuous requirement), then even the most complex classes of grammars (recursive grammars)  become learnable.\footnote{The problem of unrestricted presentations is that, for each learner, there always exists an adversarial environment that will trick the learner into converging on the wrong grammar. Vice versa, as computable processes can be enumerated, and hence a stupid learner can test increasingly many grammars and presentations and converge.} This result extends to a probabilistic scenario where the input sentences are sampled according to a statistical distribution \cite<see>{angluin_88}.

The importance of presentation boils down to the question of whether parents are being 'pedagogical' or not, i.e., whether they present language according to a \textit{curriculum} which facilitates learning\footnote{Parents may not be conscious of what they are doing: they could adjust their speech according to what they think infants hear or understand, imitate their speech, etc. By pedagogical we refer to the result, not the intent.}. %This can only be assess through a rather dense sample of realistic data. 
Importantly, such curriculum may also include phonetic aspects (e.g. articulation parameters: \citeNP{kuhl_1997}), para-linguistic aspects (e.g., communicative gestures or touch: \citeNP{csibra_2009,seidl_2015}), as well as the communication context (e.g., availability of a perceptible reference: \citeNP{sachs_1983,trueswell_2016}). %This can be assessed It is therefore important to have , which could, for some reasons or other being simplified and render less difficult the language learning task. 

To the extent that presentation matters, it is of crucial importance neither to oversimplify by assuming that parents are always pedagogical, nor to overcomplexify by assuming that there is no difference with adult-directed observations. %Estimating the nature of the curriculum from laboratory settings is problematic. Even at home, the presence of an observer may influence the parent's behavior and the communication context, making it non representative.  The only possible way to get to this variable seems to be collecting dense enough samples of parents-infants interactions in their natural environment. 

\subsubsection{How realistic does it need to be?}  We discussed three  ways in which the specifics of the input available to the learner matter greatly as to which models will succeed or fail. If one is interested in modeling infant language learning, one should therefore use inputs that are close to what infants get. How to proceed in practice?

One possible strategy would be to start simple, i.e., to work with idealized inputs generated by simple formal grammars or probabilistic models and to incrementally make them more complex and closer to real data. While this approach, pursued by formal learning theory has its merits, it faces the challenge that there is currently no known model of the variability of linguistic inputs, especially at the level of phonetics. Similarly, there is no agreed upon way of characterizing what constitutes a linguistic signal (as opposed to a non-linguistic one), nor what constitutes noise versus useful information. The particular presentation of the target language and associated contextual information that result from caretaker's communicative and pedagogic intentions has not been formally characterized. Even at the level of the syntax, the range of possible languages is not completely known, although this is perhaps the area where there are current propositions \cite<e.g.,>{jager_12}. This approach therefore runs the risk of locking researchers in a bubble universe where problems are mathematically tractable but are unrelated to that faced by infants in the real world.

A second strategy is more radical:  use actual raw data to reconstruct infant's sensory experience.  This data-driven solution is what we advocate in the reverse engineering approach: it forces to confront squarely the problem of data selection and removes the problems associated with the idealization of variability, ambiguity and mode of presentation. Importantly, the input data should not be limited to a single dataset: what we want to reverse engineer is infant's ability to learn from any mode of presentation, in any possible human language, in any modality. One practical way to address this would be to \emph{sample} from the finite although ever evolving set of attested languages, and split them into development set (to construct the algorithm) and test set (to validate it). It may be interesting to sample typologies and sociolinguistic groups in a stratified fashion to avoid overfitting the learning model to the prevalent types.% Of particular theoretical importance are the extreme case in terms of scarcity of input, multilingual diversity, or noisy 

How far should sensory reconstruction go? Obviously, it would make no sense to reconstruct stimuli outside of the sensory range of infants, or with a precision superior to their discimination abilities. Hence, input simplifications can be done according to known properties of the sensory and attentional capacities of infants. If other idealizing assumptions have to be made, at the very least, they should be explicit, and their impact on the potential oversimplification or overcomplexification of the learning problem should be discussed (as an example, see Sections \ref{section:feasibility} and \ref{section:interactive}).

\subsection{Why human-machine comparisons?}\label{section:eval}

We now turn to the apparently least controversial requirement: everybody agrees that for a modeling enterprise of any sort, a success criterion should be specified.  However, there is little agreement on which criterion to use. %In past theoretical work, the diversity is so great that it is nearly impossible to compare the different propositions across research fields (and sometimes even within field). %Prior advocates of the use of machine learning to model language acquisition have proposed a number of criteria. 

\subsubsection{Which success criterion?}
To quote a few proposals within cognitive psychology, \citeA{macwhinney_78} proposed nine criteria, \citeA{berwick_85}, nine other criteria,  \citeA{pinker_87} six criteria, \citeA{yang_02} three criteria, \citeA{frank_10} two criteria. These can be sorted into conditions about effective modeling (being able to generate a prediction), about the input (being as realistic as possible), about the end product of learning (being adult-like), about the learning trajectories, and about the plausibility of the computational mechanisms proposed.  For formal learning theorists, success is usually defined in terms of \emph{learnability in the limit} \cite{gold_1967}: a learner is said to learn a target grammar in the limit, if after a finite amount of time, his own grammar becomes equivalent to the target grammar. This definition may be difficult to apply because it does not specify an upper bound in amount of time or quantity of input required for learning (it could take a million years, see \citeNP{johnson_04}), nor does it specify an operational procedure for deciding when and how two grammars are equivalent\footnote{Two grammars are said to be (weakly) equivalent if they generate the same utterances. In the case of context free grammars, this is an undecidable problem. More generally, for many learning algorithms (e.g., neural networks), it is not clear what grammar has been learned, and therefore the success criterion cannot be applied.} .  More pragmatically, researchers in the AI/machine learning area define success in terms of the performance of their system as measured against a gold standard obtained from human adults. This may be an interesting procedure for testing the end-state of learning but is of little use for measuring learning trajectories.

We propose to replace all these criteria by a single operational principle, \textit{cognitive indistinguishability} defined in terms of cognitive tests:

\begin{displayquote}
	A human and a machine are cognitively indistinguishable with respect to a given set of cognitive tests when they yield numerically overlapping results when ran on these tests.
\end{displayquote}

Now, this definition is not sufficient in itself: it shifts  the problem of selecting a good success criterion to the problem of selecting the tests to be included in the cognitive benchmark. At least, it enables to get rid of arbitrary or aesthetic criteria (I like this model because it seems plausible, or, it uses neurons) and forces one to define operational tests to compare models. Yet, it leaves open a number of questions:  should the tests measure behavioral choices, reaction times, physiological responses, brain responses? Should they include meta- or paralinguistic tests (like the ability to detect accent, emotions, etc.)? In addition, given the range of theoretical options that have been formulated on language development (e.g., \citeNP{tomasello_03,kuhl_2000}), and disagreements on the essential properties of language (e.g.,  \citeNP{hauser_2002,evans_2009}), one would think our proposed cognitive benchmark will be difficult to come about. 

\subsubsection{How to construct a cognitive benchmark?} The benchmark that we propose to construct within the reverse engineering approach has a very specific purpose. Its aim is not to tease apart competing views of language acquisition, but to target the two developmental puzzles presented in Section 2: how do infant bootstrap onto an adult language system?  how are gradual, overlapping and resilient patterns of development possible? 

Answering these puzzles requires only to measure the state of linguistic knowledge present in the learner at any given point in development, and across the different linguistic structures (phonetic all the way to semantics and pragmatics).

This objective can be expressed in terms of the top level of Marr's hierarchy: the computational/informational level. It abstracts away from considerations about processing or neural implementation. 
This means that under such benchmark, will be considered 'cognitively indistinguishable', models of the child that have little similarity to infants psychological or brain processes (e.g. Bayesian ideal learners, artificial neural networks), so long as they have acquired the same language-specific information. Of course, one could enrich the benchmark by adding more tests that address lower levels of Marr's hierarchy (see Supplementary Section \ref{section:bioplaus} for a discussion of biological plausibility).

In addition, we propose to guide the construction of the benchmark by selecting tests that %hoice of the test by following best practices in psychophysic and psychometrics, 

satisfy three conditions: %Here, we present three conditions that such tests must satisfy to achieve our scientific objectives: 
they should be \emph{valid} (measure the construct under study as opposed to something else),  \emph{reliable} (with a good signal to noise ratio), and \emph{administrable} (to adults, children and computers alike).

The first two conditions are standard best practices in psychometrics and psychophysics
\cite<e.g.,>{gregory_2004}. Test validity refers to whether a test, both theoretically and empirically, is sensitive to the psychological construct (state or process) it is supposed to measure.  As a counterexample, the famous imitation game \citeA{turing_1950}  
tests whether machines can 'think' by
measuring how well they can appear to be humans in an on-line text-based interaction.

This test has dubious theoritical validity, as  'thinking' is not a well defined cognitive construct, but rather an underspecified folk psychology concept, and dubious empirical validity, as it is easy to fool human observers using simplistic text manipulation rules (see ELIZA, \citeNP{weizenbaum1966eliza}). Section \ref{section:task} presents Turing test replacements.

Test reliability refers to the signal to noise ratio of the measure. It can be estimated by computing the betwen-human or test-retest agreement, or by sampling over initial parameters for the machines. %Typically, test reliability is not thought to be a real issue for machines, to the extent that many algorithms are deterministic or assumed to be quite stable. Yet, it is important to assess this reliability empirically, for instance, by running the same algorithm over different samples of a large corpus. As for humans, test reliability is a very important issue, and even more so, for children and infants. 

Test administrability does not belong to standard psychometrics, but very important for comparing the performance of different systems or organisms. 

To test a human adult with most tasks, one simply provides instructions in his or her native language.

This is not directly to human infants nor to machines. In infants, a testing apparatus has to be constructed, i.e., a controlled artificial environment whereby responses to test stimuli are measured using spontaneous tendencies of the participants (preference methods, habituation methods, etc; see \citeNP{hoff_2012}, for a review).\footnote{In animals, before tests can be run, an extensive period of training using reinforcement learning is often necessary, in order for the animal to comply with the protocol. Such procedures are not possible in human infants.} As for machines, the learning algorithms are not constructed to run linguistic tests, but to optimize a particular function which may have nothing to do with the test. Therefore, they need to be supplemented with particular \emph{task interfaces} for each of the proposed tests in order to extract a response that would be equivalent to the response generated by humans.\footnote{A task interface can be viewed as a function which takes as input the internal states of the algorithm generated by the stimuli and delivers a binary or real valued response.} In all cases, administering the task should not compromise the test's validity. Biases or knowledge of the desired response has to be removed from the instructions (adults), testing apparatus (infants) and  interface (machines).

In brief, we motivated the importance of a human-machine benchmark and presented principles to construct it. The construction of the benchmark should be viewed as part of the research program itself. It should seek a common ground between competing views of language acquisition, and be periodically revised as understanding of the language competence progresses, and as new experimental protocols for language competence are established. %Therefore, the success for critera of models of language acquisition will be continuously refined. 

\section{Deep learning to the rescue? }\label{section:deep}

The combination of the first two requirements discussed above, namely, scalable computation and realistic input would have been, up to a recent period a major stumbling block for acheiving a reverse engineering approach.  Indeed, for many years, computers were struggling with language processing. It was customary in psycholinguistic courses to mock the dismal performance of automatic dictation or translation systems. All of this started to change with a paper by Hinton and colleagues on speech recognition \cite{hinton_2012}: after years in the making, neural networks were starting to perform better than the dominating technology based on probabilistic models (Gaussian Mixtures, Hidden Markov Model). A few years later, the entire speech processing pipeline has been replaced by neural networks trained end-to-end, with performance claimed to achieve human parity on a dictation task (\citeNP{xiong_2016}, but see \citeNP{saon_2017}). In the following, we very briefly review how such systems are constructed before turning on whether they could be used to inform infants language acquisition studies.

\subsection{The new AI spring}

One important characteristics of the new systems is they get rid of the specialized design features of their predecessors, and replace them with generic neural network architectures trained in large annotated corpora. Continuing with the example of speech, specialized audio features are replaced by spectrograms (some systems even work from raw audio input) and phonetic transcriptions and prononciation lexicons are eliminated: systems are trained to directly map speech to orthographic transcriptions, in an end-to-end fashion.%\footnote{End-to-end systems are followed by a statistical language model including an orthographic lexicon to clean up transcription errors, although some systems try to get rid of this to operate purely at the character level \cite{zhang_2017}.}. 

\begin{displayquote}
We do not need a phoneme dictionary, nor even the concept of a 'phoneme.'
 (\citeNP{hannun_2014}). 
\end{displayquote}

\begin{figure}
	\centering
	\includegraphics[height=5cm]{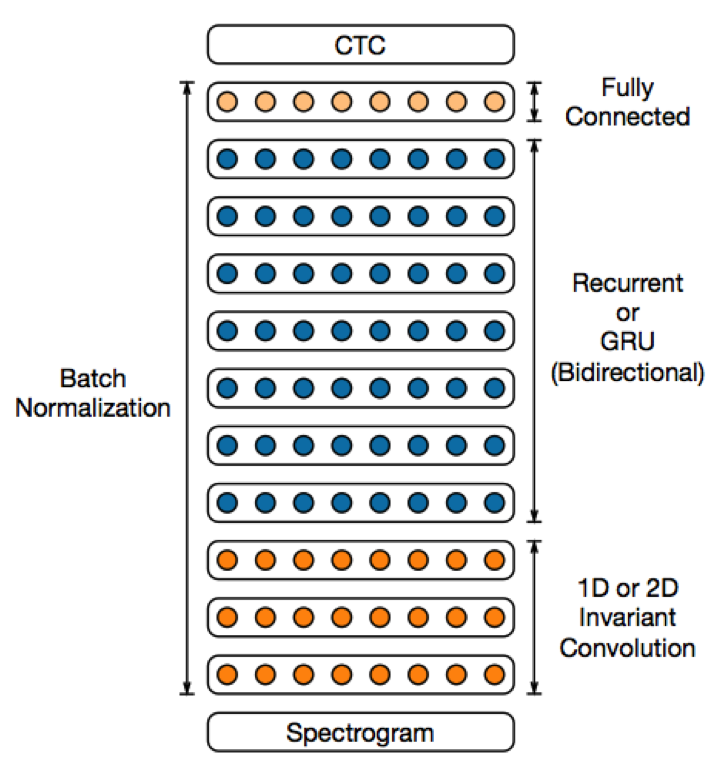}
	\caption{Network architecture for Deep Speech 2 (reprinted from Amodei et al. 2016). The input is a spectrogram, the output is a sequence of characters. Each layer incorporate particular patterns of connectivity. Convolutional layers are organized in terms of local patches sharing their connections along the time and/or frequency dimensions. Recurrent layers accumulate activations through time. Fully connected layers do not have any particular topology. Batch normalization is a process by which the activations at each layers are rescaled to be of means 0 and variance 1 over a small set of examples during training.}
	\label{fig:deepspeech2}
\end{figure}

As it turn out, the basic architectures and many core ideas are not very different from those proposed in the early days of connectionism. For instance, Figure \ref{fig:deepspeech2} shows the architecture of Deep Speech 2 \cite{amodei16} a state-of-the-art speech recognition system composed of rather classical elements popularized in the late 80's the (the multi-layer perceptron, backpropagration training, convolutional networks, recurrent networks: Rumelhart \& McClelland, 1986; \citeNP{elman_1990}).

What has changed, though, is the scale of the networks and the volume of data on which they are trained, enabled by tremendous progress in computer hardware and in mathematical optimization techniques (\citeNP{goodfellow_2016}, for an advanced introduction). As a result, neural networks have grown at a pace slightly faster than Moore's law: the speech processing network in \citeA{elman_1988} had 8000 parameters; 28 years later, Deep Speech 2 is twelve thousand times larger.  

\nocite{amodei16}

Speech is not the only area where deep learning have shaken the AI landscape: object recognition \cite{krizhevsky2012,he_2015}, language translation \cite{wu_2016,johnson_2016}, and speech synthesis \cite{oord_2016}, are all areas where neural networks %constructed with a disheartening lack of regards for cognitive theories 
have displaced by a large margin the previous state-of-the-art, while approaching human performance. This explosion of research is faciliated by the large distribution of programming frameworks (tensorflow, pytorch, dynet, mxnet, etc.), the open sourcing of datasets and state-of-the-art systems which can be downloaded pre-trained and tested on new inputs. 

These successes are generating interest for taking machine learning systems trained on large corpora as quantitative models of cognitive functions. Indeed, despite their a-priori lack of neural or biological plausibility\footnote{In fact, from their inception, neural networks have been heavily influenced by research in neuroscience and psychology, see the review by \citeA{hassabis_2017}.}, the performance of these systems show surprising convergences with biological organisms. For instance, a deep neural network trained to recognize artefacts and natural kind categories from images turn out to be good predictors of multi-unit responses of neurons in the Inferior Temporal cortex of primates \cite<e.g.,>{cadieu_deep_2014,yamins_2014}. There are also surprising divergences such as the strange way in which neural networks can be fooled by adversarial examples \cite{nguyen2014}, and limits such as their inability to perform causal reasoning or display systematic behavior \cite{lake_2016}.  This gives rise to an exciting area of research applying cognitive psychology or cognitive neuroscience methods to machine learning systems \cite[among others]{kheradpisheh_2016,cichy_2016,linzen_2016a,tsividis_2017,lau_2017}. 

The crucial question that we raise here is whether any of these algorithms could be good candidates for modeling langage acquisition? 

\subsection{Does machine learning model human learning?}

Statistical learning mechanisms have been claimed to be at the core of language acquisition \cite{saffran_2003}, so a-priori, there are reasons to be optimistic \cite{meltzoff_2009}. However, there is a fundamental gap between what this term means in cognitive studies and how it is used in machine learning. The big difference is that in machine learning, statistical techniques are used as a convenient way to construct systems, not as models of human acquisition processes. 

Interpreted cognitively, machine learning procedures would correspond to a caricature of 19$^{th}$ century schooling: the learner, initially, a kind of tabula rasa, is relentlessly fed with inputs paired with desired responses, which are annotations of the input provided by a human supervisor. The drill is repeated until the learner gets it right. %response is correct. 

This setup is called \textit{supervised learning}, because for a given input there is only one correct answer. %, patiently provided by the supervisor during the training phase. 
As an example, in speech recognition, the system is  trained to associate a speech utterance with it's written transcription.  In natural language processing tasks, the system is presented with sequences of words (in ortographic format) as input, and trained to associate each word to a part-of-speech, a semantic role tag, or a co-reference in the text, and so on. % In dialogue tasks, systems are trained to associate to each turn (sentence),  speech act. 
This differs in how infants learn language in two important ways. 

First, children do not learn their first language by being asked to associate sensory inputs with linguistic tags. Long before they are even exposed to linguistic tags by going to school and learn to read and write, they have acquired what amounts to a fully functional speech recognition and language processing system. They have done so on the basis of sensory input alone, and if there are supervisatory signals from the adults, these are neither unambiguous nor systematic. This moves the problem of language learning in the area of \textit{unsupervised} or \textit{weakly supervised} machine learning: to an input, there is no unique desired output, but rather a probabilistic distribution of outcomes (with relatively unfrequent rewards or punishments)\footnote{This raises the issue about what is the internal reward for the infant which pushes him or her to acquire language. A drive for learning statistical patterns? A drive to interact with others in his or her group?}. 
 
The second difference is in the sheer amount of data required by artificial systems compared to infants. For instance, the Deep Speech 2 system described above is trained with over 10000 hours of transcribed speech (plus a few billion words worth of text to provide top-down language statistics).  In comparison, a four-year-old child, who admittedly has functional speech recognition abilities, is being spoken to for a total amount varying between 700h and 4000h  (corresponding to 8 and 44M words, respectively), depending on the language community (for estimates, see Supplementary Section \ref{section:estimate}). This means that Deep Speech 2 requires around 14 times more speech, and 240 times more words than what a four-year old Mayan child get. A recent time allocation study in the Tsimane community \cite{cristia:2017}  shows that the amount of child directed input may even be lower than the Maya yet by a factor of 3 (less than one minute of speech per waking hour). This shows that the human infant is equipped with a learning algorithm which enables him or her to learn language with very scarse data. %The Tsimane are a forager society which is closer to the kinds of societies in which humans spent most of their evolutionnary history than urban industrialized societies. This shows that the learning techniques used currently in machine learning are much less data efficient than human infants are. 

\subsection{Summing up.}

Machine learning has made progress to the point that 'cognitive services' (speech recognition, automatic translation, object and face recognition, etc.) are incorporated in everyday life applications. This means that one of the major road block for the reverse engineering approach, i.e. the feasibility of building language processing systems that can deal with realistic input at scale is now lifted.  Instead of being locked with simplified data or toy problems, for the fist time, it becomes possible to address the bootstrapping problem in its full complexity, and derive quantitative developmental predictions along the way.

Still, there are challenges ahead; current machine learning systems fail to provide models of infant acquisition, not because they discard or simplify the input, but because they use too much of it, both in sheer quantity and in adding extra inputs that the infant could not possibly get (linguistic labels). What needs to be done, therefore is to adapt some of the existing algorithms or construct new ones, so that they can learn with as few data as infants do.  How far are we?% from this? This is what we examine next. 

\section{The road ahead}

\label{section:feasibility}

We now turn to the feasibility of the reverse engineering approach as applied to early language development. To do so, we limit ourselves to the following simplifying assumption: 

\begin{displayquote}
The total input available to a particular child provides enough information to acquire the grammar of the language present in the environment. 
\end{displayquote}

\begin{figure}
	\centering
	\includegraphics[height=3.7cm]{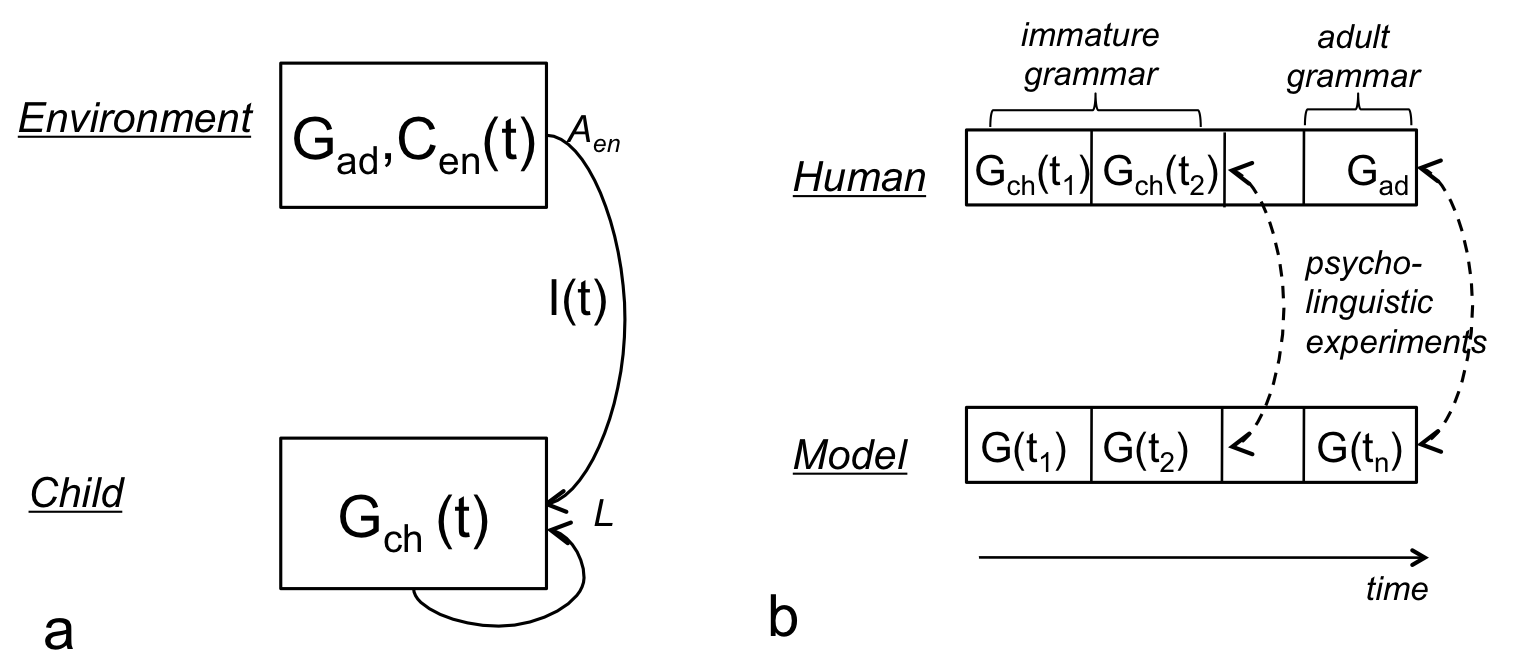}
	\caption{a. The (simplified) learning scenario: The Child's internal state is a grammar $G_{ch}(t)$ that can be updated through the learning function $L$ based on input $I(t)$. The environment's internal state is a constant adult grammar $G_{ad}$ and a variable context $C_{en}$, which produces the input to the child. b. Method to test the empirical adequacy of the model by comparing the outcome of psycholinguistic experiments with that of children and adults.}
	\label{fig:relaf1}
\end{figure}

This may seem reasonable, but it essentially puts us in the open loop situation described in Figure  \ref{fig:relaf1}),  where the environment delivers a fixed curriculum of inputs (utterances and their sensory contexts) and the learner recovers the grammar that generated the utterances. In this situation, the output of the child is not modeled, and the environment does not modify its inputs according to her behavior or inferred internal states. This input-driven idealization may overestimate the difficulty of the task compared to a more realistic close-loop scenario. We think however, that it is useful to study the input-driven scenario in its own sake, as it gives an estimate of what can be learned in the worse case scenario where parents have other priorities than optimizing their children's language learning.  

We examine how this simplifying assumption can be relaxed in Supplementary Section \ref{section:interactive}

Within this scenario, we claim that recent advances in AI and big data now make the reverse engineering roadmap actionnable. We discuss current avenues of research and the challenges that need to be met.  Following our three requirements, we review, in turn, the feasibility of constructing systems that can learn without expert labels, the collection of large realistic dataset, and the establishment of human-machine benchmarks, and illustrate it with a selection of recent work.

\subsection{Unsupervised / weakly supervised algorithms}

Bringing machine learning to bear to language development requires to construct systems that discover linguistic structure with little or no expert supervision.  This is
obviously more difficult than learning to associate inputs to linguistic labels. Here, the learner has to discover its own labels given the input.  This class of machine learning problems is unfortunately less well studied and understood than supervised learning, but is an expanding field of research in machine learning. Two main, non exclusive, ideas are being explored to address this challenge.

\subsubsection{Inductive biases.}  The first idea is to build into the learner %\textit{inductive biases}, which incorporates
 prior knowledge about the underlying nature of the data, so that generalization can be made with few or noisy datapoints. % and constrains the way in which the input data is interpreted. 
With strong prior knowledge, some logically impossible learning problems become easily solvable.\footnote{One good illustration is the following: can you tell the colors of 1000 balls in an urn by just selecting one ball? The task is impossible without any prior knowledge about the distribution of colors in the urn, but very easy if you know that all the balls have the same color.} %Similarly, if the learner is well prepared to specific problems, it may learn with fewer and weaker datapoints than a very generic learning. 
Some models of the acquisition of syntax mentioned in Section 3.1 favor very strong priors, where the only thing to learn (besides the meaning of words) is a small number of syntactic binary parameters. The learning problem becomes so constrained that a single sentence (called a trigger) may be sufficient to decide a parameter's value \cite{gibson_1994,sakas_2012}. The notion of inductive biases can be formulated elegantly using Bayesian graphical models \cite{pearl_probabilistic_1997,koller_2009}. In these models, prior knowledge is specified as probability distributions over the model's parameters, which are updated for each new input (see \citeNP{gershman_2015} for a general presentation). 

For the purpose of illustration, let us revisit the discovery phonetic categories from continuous speech. We have mentionned previously that generic clustering algorithms fail to learn phonemes, because of a mismatch between what clustering algorithms expect (relatively well delimited clusters) and what the data consists in (a complicated gesture unfolding in time). %Within a Bayesian framework it is possible to construct a better suited model.  
\citeA{lee_2012} proposed a Bayesian graphical model, where phonemes are defined as sequences of three acoustic states (schematically, a state for the beginning, the central part and the end of the phoneme). Each state is modeled as a mixture of 8 Gaussians in the space of acoustic parameters (MFCCs, a representation derived from spectrograms). Phoneme durations are also controlled through a binary boundary variable (modelled with a poisson distribution), and the number of phonemes is specified by a Dirichlet prior, which expects the distribution of phonemes to follow a power law (a few phonemes are used often, many phonemes are used rarely). Far from being a general purpose clustering algorithm, the algorihm of Lee \& Glass uses language-universal information about the phonemes, (their shape, their duration, their frequency) to specify a model that will be inductively biased to discover this kind of structure in the data. Bayesian probabilistic models are also used in natural language processing to infer syntactic structures from raw data without supervision
\cite{liang_2011,kwiatkowski_2012}. Some of these models have been recently used on child directed input (CHILDES transcripts) to account for developmental results \cite{abend_2017}. 

The challenge with these types of models is that the optimization of the parameters is very computationally intensive, which becomes prohibitive for large models and/or large datasets. For instance, the Lee \& Glass model has only been applied to a relatively small corpus of read speech (TIMIT), and the \citeA{abend_2017} model on textual input.  Current research is devoted to develop efficient approximations of these algorithms to deploy them in more naturalistic datasets (see for instance \citeNP{ondel_2016} for a scalable reimplementation of Lee \& Glass).

\subsubsection{Synergies.} Here, the idea is that the different components 
of language being interdependant, it may help to jointly learn these components rather than to learn them separately. This is actually turning the bootstrapping problem on its head: instead of being a liability, the codependancies between linguistic components become an asset. Of course, it is an empirical issue as to whether joint learning between any two language components is always more successful than separate learning.  The existence of synergies has been documented using Bayesian models between phonemes and words inventories \cite{feldman_2011}, syllables and words segmentation \cite{johnson_2008}, referential intentions and word meanings \cite{frank_2009}.

\begin{figure}
	\centering
	\includegraphics[height=8cm]{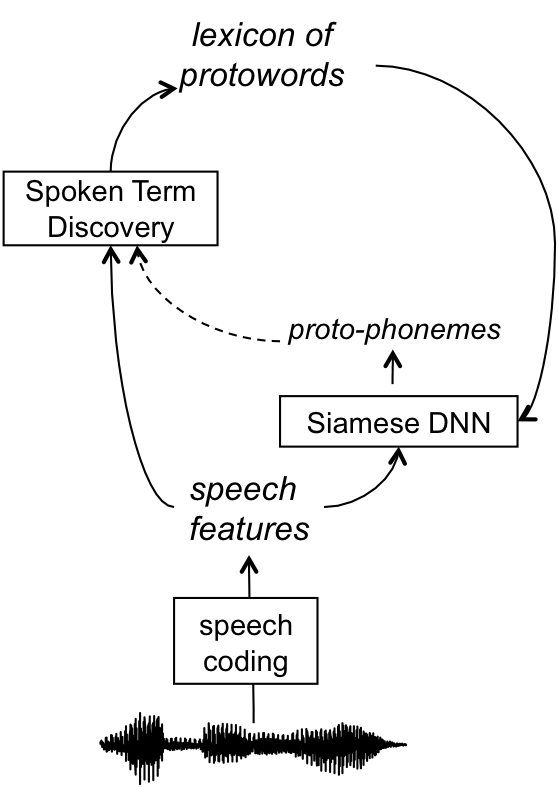}
	\caption{Architecture illustrating a top-down synergy between learning phonemes and words. Auditory spectrograms (speech features) are computed from the raw speech signal. Then, protowords are extracted using Spoken Term Discovery; these words are then used to learn a more invariant speech representation using discriminative learning in a siamese Deep Neural Network architecture (from Thiolliere et al., 2015).}
	\label{fig:monsterbeta}
\end{figure}

The existence of  synergies can be leveraged in models other than Bayesian ones, including deep learning or  more algorithmic speech engineering systems. 
For instance, returning to the issue of phonetic learning, several lines of research indicate that words could help the discovery of subword units \cite{swingley_2009, thiessen_2007}), and that even an imperfect, automatically discovered proto-lexicon can help \cite{martin_2012, fourtassi2014c}. The model described in Figure \ref{fig:monsterbeta} implements this idea. It consists in a word discovery system which extracts similar segments of speech across a large corpus. The discovered segments constitute a proto-lexicon of acoustic word forms \cite{jansen_2013b}, which are then used to train a neural network in a discriminative fashion. The resulting output of the network is a representation of speech sound which is much more invariant to a change in talker than the original spectral representation on which the system started with \cite{thiolliere_2015}.  
In a similar spirit, \citeA{harwath_2016} and \citeA{harwath_2017} showed that by training a neural network to associate an image with a speech input corresponding to a short description of this image, the network develops phone-like and word-like intermediate representations for speech.

\begin{figure}
	\centering
	\includegraphics[height=5cm]{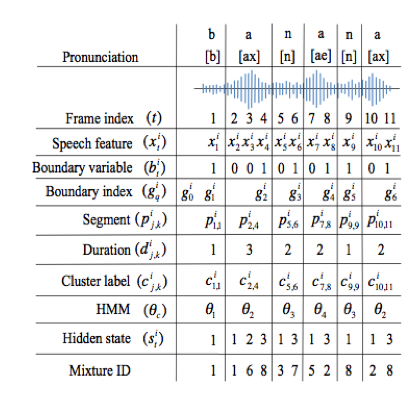}
	\caption{Outline of a clustering algorithm with a hierarchical generative architecture for learning phonemes from raw speech (reprinted from Lee, \& Glass, 2012). The model is provided with speech described by speech features (3$^{rd}$ line), and infers all of the other parameters (4${th}$ line downwards: boundary variables, segment identify and duration, states of the hidden markov model and parameters of the Gaussian mixtures) from the data according to the hierarchical model by sampling in the space of possible values for these parameters.}
	\label{fig:glass}
\end{figure}

In brief, even though unsupervised/weakly supervised learning is difficult, there is a growing interest within machine learning for the study of such algorithms, as shown by special sessions on this topic in machine learning conferences, and the organization of challenges involving laboratories in cognitive science and speech technology communities (e.g. the zero resource speech challenge, \citeNP{versteegh_2015,dunbar2017}).

\subsection{Large scale data collection in the wild}\label{section:privacy}

A large number of datasets across languages have been collected and organized into repositories that have proved immensely useful to the research community. One prominent example of this is the CHILDES repository \cite{macwhinney2000childes}, which has enabled more than 5000 research papers (according to a google scholar search as of 2016). These datasets, however, contain only relatively sparse datapoints (a few hours per infants).
Perhaps the most ambitious large scale and dense data collection effort to date is the Speechome project \cite{roy_2009}, where video and audio equipment was installed in each room of an apartment, recording 3 years' worth of data around one infant. This pioneering work illustrates several key technological, analysis and ethical issues that arise in 'ecological' data collection.

Regarding technological issues, the falling costs in digital sensors and data storage make it feasible to duplicate Speechome-like projects across many languages. More challenging is the fact that to be usable for modeling, the captured should enable the reconstruction of infant's sensory experience from a first person point of view.  Already, relatively inexpensive out-of-the box wearable technology can go some way in that direction. Miniaturized recorders (see for instance the LENA system, \citeNP{xu_08}) enable recording the infant's sound environment for a full day at a time, even outside  home, and will become more and more usable as microphone array and advanced signal processing enable source reconstruction even in noisy environment. Proximity and accelerometor sensors can be used to categorize activities \cite{sangwan_2015}; 'life logging' wearable devices capture images every few seconds and help to reconstruct the context of speech interactions \cite{casillas_2016}. Head-mounted cameras can help to reconstruct infant's field of view \cite{smith_2015}.  Upcoming progress in the miniaturization of 3D sensors would enable to go further in the reconstruction of infant's visual experience.  %Recent progress in 3D reconstruction, especially when using multi-view and/or depth sensors make it possible to go further in sensory reconstruction \cite<e.g.,>{mustafa_2016} although this has not yet been done with infant data. 

Regarding analysis issues, the challenge it to supplement raw data with reliable linguistic/high level annotations. Manual annotations are too costly to scale up to  large and dense datasets. In the Speechome corpus, more than 3000 hours of speech have been transcribed, wich represents only a fraction of the total 140000 hours of audio recordings \cite{roy_2015}. The recent breakthroughs in machine learning discussed in Section \ref{section:deep} (speech recognition: \citeNP{amodei16}; object recognition: \citeNP{girshick_2016}; action recognition: \citeNP{rahmani_2016}; emotion recognition: \citeNP{kahou_2015}) will enable the semi-automatic annotations of large amounts of data.

As for ethical issues, % raised by the need to make this data accessible to the research community. There is a tension between 
the main challenge is to find a point of equilibrium between the requirement of sharability and open scientific data, and the need of protecting the privacy of the familie's personal  data. Up to now, the response of the scientific community has been dichotomous: either make everything public (as in the open access repositories like CHILDES, \citeNP{macwhinney2000childes}), or completely close off the corpora to anybody outside the institution that has recorded the data (as in the Riken corpus, \citeNP{mazuka_2006}, or the Speechome corpus \citeNP{roy_2009}). Neither solutions are acceptable. 

Alternative strategies are being considered by the research community. The Homebank repository contains raw and transcribed audio, with a restricted case by case access to researchers (\citeNP{vandam_2016}, \url{http://homebank.talkbank.org}). Databrary has a similarly organized system for the secure storage of large sets of video recordings of developemental data (\citeNP{gilmore_2017}, \url{https://nyu.databrary.org}). Progress in cryptographic techniques would make it possible to envision preserving privacy while enabling more open exploitation of the data. For instance, the raw data could be locked on secure servers, thereby remaining accessible and revokable by the infants' families. Researchers' access would be restricted to anonymized meta-data or aggregate results extracted by automatic annotation algorithms. %\emph{Differential privacy} techniques enable outside participants to make queries on databases while providing a level of guarantee on the amount of private information that can be extracted \cite{dwork2006differential}. 
The specifics of such a new type of linguistic data repository would have to be worked out before dense speech and video home recordings can become a mainstream tool for infant research. 

In brief, large scale data collection of infant data is within reach and is under in a number of research projects (see \url{www.darcle.org}), although it's exploitation in an open source format requires specific developments in privacy-preserving storage and computing infrastructures.  

\subsection{Cognitive benchmarking of language acquisition}\label{section:task}

\newcommand{\pb}[1]{\parbox[t][][t]{.99\linewidth}{#1}}

\begin{table*}[]
	\small
\centering
\caption{Example of tasks that could be used for a Cognitive Benchmark.}
\label{tab:task}

\begin{tabular}{p{0.22\linewidth}p{0.12\linewidth}p{0.36\linewidth}p{0.31\linewidth}}

\hline
Task description in human adults	&Linguistic level & Equivalent task in children	&Equivalent task in machines\\

\hline

\pb{Well-formedness judgement\\ \emph{does utterance S sound good?}}&\pb{phonetic, prosody, phonology, morphology, syntax}& \pb{preferential looking (9-month-olds: \citeNP{jusczyk_1997}), acceptability judgment (2-year-olds: de Villiers and de Villiers,  1972; Gleitman, Gleitman, and Shipley, 1972)}& \pb{reconstruction error \cite{allen_1999}, probability \cite{hayes_2008}, mean or min log probability \cite{clark_2013}}\\	
\\	
\pb{Same-Different judgment\\ \emph{is X the same sound / word / meaning as Y?}}&\pb{phonetic, phonology, semantics} &\pb{habituation / deshabituation (newborns, 4-month-olds: \citeNP{eimas_1971,bertoncini_1987}), oddball (3-month-olds: \citeNP{dehaene_1994})} 	&\pb{AX/ABX discrimination \cite{carlin_2011,schatz_2013}, cosine similarity \cite{landauer_1997}}\\
\\

\pb{Part-Whole judgment\\ \emph{is word X part of sentence S?}}&phonology, morphology & Word spotting (8-month-olds: \citeNP{jusczyk_1999})& spoken web search \cite{fiscus_2007} \\
\\

\pb{Reference judgment\\ \emph{does word X (in sent S) refer to meaning M?}}&semantics, pragmatics&\pb{intermodal preferential looking (16-month-olds: \citeNP{golinkoff_1987}), picture-word matching (11-month-olds: \citeNP{thomas_1981})} &	picture/video captioning \cite<e.g., >{devlin_2015}, Winograd's schemas \cite{levesque_2011}\\
\\
\pb{Truth/Entailment judgment\\ \emph{is sent S true (in context C)?}}&semantics&\pb{Truth Judgment Task (3-year-olds: \citeNP{abrams_1978,lidz_2002})}& visual question answering \cite{antol_2015}	\\
\\
\pb{Felicity judgement\\ \emph{would people say S to mean M (in context C)}?}& pragmatics&\pb{Ternary reward task (5-year-olds: \citeNP{katsos_2011}), Felicity judgment task (5 years olds: \citeNP{foppolo_2012}).} &?	\\

\hline
\end{tabular}
\end{table*}

\nocite{devilier_72,gleitman_72}

Our final requirement, the construction of a cognitive benchmark for language processing, can draw from work in linguistics and psycholinguistics. 

On can indeed find relatively easy-to-administer, valid and reliable tests of the main components of linguistic competence in perception/comprehension (see Table \ref{tab:task}). These tests are easy to administer because they are conceptually simple and can be administered to naive participants; most of them are of two kinds: goodness judgments (say whether a sequence of sound, a sentence, or a piece of discourse, is 'acceptable', or 'weird') and matching judgments (say whether two words mean the same thing or whether an utterance is true of a given situation, which can be described in language, picture or other means). As for validity, (psycho)linguistic tests often use a \emph{minimal set design} where one linguistic construct is manipulated while every other variable is kept constant (for instance: 'the dog eats the cat' and 'the eats dog the cat' contain the same words, but one sequence is syntactically correct, the other not).  Regarding test reliability, as it turns out, many linguistic tests are quite reliable, as 97\% of the results of a grammaticality judgment from textbooks are replicable using on-line experiments \cite{sprouse_2013}\footnote{Of course, even in simple psychophysical tasks, humans can be affected by many other factors like attention, fatigue, learning or habituation to stimuli or regularities in stimulus presentations, etc. Methods try to minimize but never totally suceed in neutralizing these effects.}.

Given the simplicity of these tasks, it is relatively straightforward to apply them to machines. Indeed, matching judgments between stimulus A and stimulus B can be derived by extracting from the machine the representations triggered by stimulus A and B, and compute a \textit{similarity score} between these two representations. Goodness judgments are perhaps more tricky; they can easily be done by generative algorithms that assign a \textit{probability score}, a \textit{reconstruction error}, or a \textit{prediction error} to individual stimuli. 
As seen in Table \ref{tab:task}, some of these tests are already being used quite standardly in the evaluation of unsupervised learning systems, in particular, in the evaluation of phonetic and semantic levels while for others they are less widespread.\footnote{Regarding the evaluation of word discovery systems, see the proposition by \citeA{ludusan_14} but see \citeA{pearl_2016} for a counter proposal and a discussion in \citeA{dupoux_2016}.}

The challenge comes from the applicability of these tests to infants and children. As seen in Table \ref{tab:task}, there are considerable variations on the age at which different linguistic levels have been tested in children, and generally, the younger the child, the more difficult it is to construct reliable tests. %The replicability crisis (see \citeNP{ioannidis2012science}; Open Science Collaboration, 2015\nocite{open_2015}) has barely hit developmental psychology yet because there are so few replications in the first place (although, see the Many Babies project, \citeNP{frank_2016}).  
Addressing this challenge would require improving substantially the signal-to-noise of some of these techniques. %Existing meta-analyses highlight large differences in effect sizes across experimental methods (community-augmented meta-analyses: ),  which point to ways to improve them.
There is also the possibility to increase the number of participants through community-augmented meta-analyses (\citeNP{tsuji2014community}; see also \url{http://metalab.stanford.edu/}), collaborative testing \cite{frank_2017}, % as in genome-wide association studies,  where low power requires a consortium to run very large number of participants (e.g., around 200,000 participants in \citeNP{ehret_2011}) 
or remotely run experiments (Shultz, 2014, \url{https://lookit.mit.edu/}; Izard, 2016,  \url{https://www.mybabylab.fr}).

Before the full reverse engineering roadmap has been put into place, it is already possible to test specific predictions using existing techniques. One can use the patterns of errors made by computational models when run on infant input data to generate new predictions. The reasoning is that these errors should not be viewed as 'bugs', but rather signatures of intrinsic computational difficulties that may also be faced by infants. For instance, even very good word discovery algorithms make systematic segmentation errors: under-segmentations for frequent pairs of words (like "readit" instead of "read"+"it") or over-segmentations  ("butter"+"fly" instead of "butterfly") \cite<see>{peters_1983}.%  Instead of viewing these errors are inadequacies of the models, one could view them as predictions: if it is computationally difficult to segment certain words, it is reasonable to expect that infants would make the same errors, at least at an age where the assumptions of the models are approximately correct.  These 'errors' could then be presented into infants and tested for recognition. If infants are making the same errors as the proposed mechanism, this would count as evidence in favor of this mechanism.  

\citeA{ngon_2013} showed that it is possible to use the preferential listening paradigm in eleven month infants to probe for signature mis-segmentations. Deriving predictions from a very simple model of word discovery (an ngram model) run on a CHILDES corpus, she constructed a set of otherwise matched frequent versus unfrequent mis-segmentations. Eleven month olds preferred to listen the frequent mis-segmentations, and did not distinguish them from real words of the same frequency. \citeA{larsen_2017} found that it was possible to compare the outcome of different segmentation algorithms in measuring their ability to predict vocabulary acquisition as measured by parental report.

In brief, while a cognitive benchmark can be established, and it is already possible to test in infants some predictions of computational models, large scale model comparison will require progress in developmental experimental methods.

\section{Conclusions}\label{section:conclusion}

During their first years of life, infants learn a vast array of cognitive competences at an amazing speed; studying this development is a major scientific challenge for cognitive science in that it requires the cooperation of a wide variety of approaches and methods. Here, we proposed to add to the existing arsenal of experimental and theoretical methods the reverse engineering approach, which consists in building an effective system that mimics infant's achievements. The idea of constructing an effective system that mimics an object in order to gain more knowledge about that object is of course a very general one, which can be applied beyond language (for instance, in the modeling of the acquisition of naive physics or naive psychology) and even beyond development. % Related work exist in the area of computational neuroscience, which attempts to use machine learning architectures (deep learning) and bring it to bear to the analysis of neural representations for visual  inputs \cite{cadieu_deep_2014,isik_2016,leibo_2015}. The computational rationality framework uses bayesian modeling to bring together the field of Artificial Intelligence and studies of human abilities like reasoning or decision making \cite{gershman_2015}.

We have defined three methodological requirements for this combined approach to work: constructing a computational system at scale (which implies 'de-supervising' machine learning systems to turn them into models of infant learning), using realistic data as input (which implies setting up sharable and privately safe repositories of dense reconstructions of the sensory experience of many infants), and assessing success by running tests derived from linguistics on both humans and machines (which implies setting up benchmarks of cognitive and linguistic tests). We've showed that even before these challenges are all met, such an approach can help challenging verbal theories, help characterize the learning consequences of different kinds of  inputs available to infant across cultures, and suggesting new empirical tests.

Before closing, let us note that the reverse engineering approach we propose does \emph{not} endorse a particular model, theory or view of language acquisition. For instance, it does \emph{not} take a position on the rationalist versus empiricist debate (e.g., Chomsky, 1965, vs. Harman, 1967). Our proposal is more of a methodological one: it specifies what needs to be done such that the machine learning tools can be used to address scientific questions that are relevant for such a debate. It strives at constructing at least one effective model that can learn language. Any such model will both have an initial architecture (nature), and feed on real data (nurture). It is only through the comparison of several such models that it will be possible to assess the \emph{minimal} amount of information that the initial architecture has to have, in order to perform well. Such a comparison would give a quantitative estimate of the number of bits required in the genome to construct this architecture, and therefore the relative weight of these two sources of information. In other words, our roadmap does not start off with a given position on the rationalist/empiricist debate, rather, a position in this debate will be an outcome of this enterprise.

\section*{Acknowledgments}
This paper would not have come to light without numerous inspiring discussions with Paul Smolensky and Alex Cristia.  It also benefitted from insightful comments by Paul Bloom, Emmanuel Chemla, Ewan Dunbar, Michael Frank, Giorgio Magri, Steven Pinker, Thomas Schatz, Gabriel Synnaeve, the members of the Cognitive Machine Learning team of the Laboratoire de Sciences Cognitives et Psycholinguistique, and three anonymous Cognition reviewers. This work was supported by the European Research Council (ERC-2011-AdG-295810 BOOTPHON), the Agence Nationale pour la Recherche (ANR-10-LABX-0087 IEC, ANR-10-IDEX-0001-02 PSL*), the Fondation de France, the Ecole de Neurosciences de Paris, and the Region Ile de France (DIM cerveau et pens\'ee).

\section*{Appendix: Supplementary Materials}

\renewcommand\thesubsection{S\arabic{subsection}.}    
\renewcommand\thesection{S\arabic{section}.}    
\renewcommand\thetable{S\arabic{table}}    
\renewcommand\thefigure{S\arabic{figure}}

\setcounter{secnumdepth}{2} % in order to number the sections and subsections
\setcounter{section}{0}  
\setcounter{table}{0}  
\setcounter{figure}{0}  

\section{Estimate of input to child}\label{section:estimate}

\begin{table*}[]
	\centering
	\caption{Four studies used to estimate infant's speech input }
	\label{tab:inputstudies}
	\begin{tabular}{p{0.1\linewidth}p{0.25\linewidth} p{0.25\linewidth} p{0.25\linewidth}}
		\hline
		study& reference & mode of acquisition;age & population\\
		\hline
		
		H\&R &\raggedright\citeA{hart_95} &\raggedright
		observer, 1h every month; 12-36 months & urban high, mid \& low SES, English  \\
		S\&G &\raggedright\citeA{shneidman_12} &\raggedright
		observer, 1h every month; 12-36 months & urban high SES, English \& rural low SES, Maya \\ 
		W\&F &\raggedright\citeA{weisleder_13} &\raggedright
		daylong recording; 19 months            & low SES, Spanish            \\
		VdW\ &\raggedright\citeA{van_de_weijer_02} &\raggedright
		daylong recording; 6-9 months           & high SES, Dutch            \\
		\hline
	\end{tabular}
\end{table*}

\newcommand{\ts}{\textsuperscript}

\begin{table*}[]

	\centering
	\caption{Estimates of yearly input,  in total, and restricted to Child Directed Speech (CDS) , in number of hours and words (millions) per year in four studies (see the references in Table \ref{tab:inputstudies}) as a function of sociolinguistic group (SES: Socio Economic Status). The numbers between brackets provide the range [min, max] of these numbers across families. \ts{t}~uses a wake time estimate of 9 hours per day. \ts{w}~uses a word duration estimate of 400ms. \ts{c}~uses S\&G's estimate of \%CDS for high SES. \ts{d}~uses W\&F's estimate of \%CDS for low SES. \ts{m}~uses H{\&}R's MLU's estimates (according to SES). %Note: the data of SALG is currently not available.
		}
	\label{tab:input}
\resizebox{\textwidth}{!}{
	\begin{tabular}{llllllllll}
		\hline
		& \multicolumn{4}{c}{Yearly total} && \multicolumn{4}{c}{Yearly CDS} \\
		\cline{2-5} \cline{7-10} 
		& \multicolumn{2}{c}{Hours}   & \multicolumn{2}{c}{Words (M)}  && \multicolumn{2}{c}{Hours} & \multicolumn{2}{c}{Words (M)}  \\
		\hline
		\multicolumn{10}{c}{Urban, high SES}                                                \\
		H{\&}R (N=13)\ts{t} &1221\ts{w,c} &[578,1987] & 11.0\ts{c} &[5.20, 17.9]&&786\ts{w}   &[372, 1279]& 7.07      &[3.35, 11.5]  \\
		S\&G (N=6)\ts{t} &2023\ts{w,m} &[1243, 2858]& 18.2\ts{m} &[11.2, 25.7]&&1223\ts{w,m}&[853, 1574]& 11.0\ts{m}&[7.7, 14.2] \\
		VdW (N=1)        & 931         &             & 9.28       &            &&140         &            & 1.39      & \\
		\\
		\multicolumn{10}{c}{Urban, low SES}                                                 \\
		H{\&}R (N=6)\ts{t}   & 363\ts{w,d} &[136, 558] & 3.26\ts{d}&[1.22, 5.02] && 225\ts{w} &[84, 346] & 2.02 &[0.76., 3.11]\\
		W{\&}F (N=29)\ts{t} & 363\ts{w}   &[52, 1049]  & 3.27        &[0.46., 9.44]&& 225\ts{w} &[32, 650] & 2.03 &[0.29, 5.85]\\
		\\
		\multicolumn{10}{c}{Rural, low SES}                                                               \\
		S\&G (N=6)\ts{t} & 503\ts{w,m} &[365, 640] & 4.53\ts{m}&[3.28, 5.76]&& 234\ts{w,m}&[132, 322] & 2.10\ts{m}&[1.19, 2.90] \\
		%S\&G (N=X, 13 month only)\ts{t} & 348\ts{w,m}& 3.14\ts{m}& 69.7\ts{w,m}& 0.63\ts{m}  \\
		\\
		%\multicolumn{5}{c}{Urban, Worst case scenario estimate}                            \\
		%H\&R (1\ts{st} decile, 12-36mo) & 106      & 0.95       & 65.7     & 0.59          \\
		%W\&F (1\ts{st} decile, 12-36mo) & 54.6     & 0.49       & 27.7     & 0.25          \\
		\hline
	\end{tabular}
}

\end{table*}
In this section, we describe the way in which we estimated the amount and variability of speech input to infants.  We are mainly interested in the number of hours and number of words, since these are two common metrics used in automatical speech recognition and natural language processing. We therefore use these metrics when they were available in the original data, and estimate them otherwise. 

Table \ref{tab:input} lists the sample of four studies that we have included in our survey, which incorporate large variations in languages and cultures. \citeA[H\&R]{hart_95} studied English speaking infants splitted into three groups according to the Socio-Economic Status (SES) of the familly. In our analysis, we only include the two extreme groups (N=13 and 6, respectively). \citeA[S\&G]{shneidman_12} studied two groups, one rural Mayan speaking community (N=6), one English speaking urban community in the USA (N=6). \citeA[W\&F]{weisleder_13} studied one group of low SES Spanish speaking familly in the USA (N=29). Finally, \citeA[VdW]{van_de_weijer_02} extensively measured one Dutch speaking child in the Netherlands.

One methodological problem is that the four studies reported different kinds of metrics (H\&R: number of words and utterances, S\&G: number of utterances, W\&F: number of words, and VdW: number of hours, words and utterances). In order to compare them, one has therefore to estimate how to convert one metric into another, which requires possibly incorred assumptions about the conversion parameters. They should therefore be taken with a  large grain of salt, and are subject to revision when more precise data comes along.

Table \ref{tab:inputstudies} lists the results and indicate the value of the conversion factor that we used.  To compute the total number of hours per year, we used a waking time estimate of 9h for all of the studies except VdW which directly estimated  speaking time per day. To convert number for words into hours, we used an estimate of word duration of 400ms. This is compatible with the numbers reported by VdW. To convert between number of utterances and number of words, we used an SES-dependant estimate of Mean Utterance Length of 4.43 for high SES and 3.47 for low SES (from H\&R). Finally, to estimate the total amount of speech heard by infants, we used a proportion of Child Directed Input of 64\% for high SES (for S\&G) and of 62\% for low SES (from W\&F). To see an updated version of this analysis including a new population of forager-farmers, see \cite{cristia:2017}.

\section{A biological plausibility requirement?}\label{section:bioplaus}

In this section, we briefly discuss one issue which often comes up when computational systems are used as models of human processing: the issue of \emph{biological plausibility}. By this, we mean that the hypothetical algorithm be compatible with what we know about the biological systems that underlie these computations in human infants/adults. 

While this constraint is perfectly reasonable, we argue that it is difficult to apply to the modeling of early language acquisition for the following reasons: First, the computational power of a human brain is currently unknown. Current supercomputers can simulate at a synapse level only a fraction of a brain and several orders of magnitude slower than real time \cite{kunkel_2014}. If this is so, all computational models run in 2016 are still massively underpowered compared to a child's brain.  Second, a particular algorithm may appear to be too complex for the brain, but a different version performing the same function will not.  For instance, some word segmentation algorithms require a procedure called Gibbs sampling, which, in theory, require an infinite number of time steps to converge. This would seem to discredit the algorithm alltogether. Yet, it turns out that a truncated version of this algorithm running in finite time works reasonably well. Similarly, algorithms that require a lot of time steps can be rewritten into algorthms that require less steps and more memory. This makes a priori claims of biological plausibility difficult to make.

Still, biological plausibility can place some theoretical bounds on \emph{system complexity at the initial state}. Indeed, the initial state is constructed on the basis of the human genome plus prenatal interactions with the environment. This allows to rule out, for instance, a 100\% nativist acquisition model that would pre-compile a state-of-the-art language understanding systems for all of the existing 6000 or more languages on the planet, plus a mechanism for selecting the most probable one given the input.\footnote{The reason such system would not be biologically realizable is that the parameters of a state-of-the-art phoneme recognition system for a single of these languages already require 10 times more memory storage than what is available in the fraction of the genome that differentiate humans from apes. A DNN-based phone recognizer has typically more than 200M parameters, which barring ways to compress the information, takes 400Mbytes. The human-specific genome is 5\% of 3.2Gbase, which boils down to only 40Mbytes.}

Apart from this rather extreme case, biological plausibility may not affect much of the reverse engineering approach until more is known about the computational capacity of the brain. Yet, it is compatible with our approach, since as soon as diagnostic tests of language computation in the brain are available, they could be added to the cognitive benchmark, as defined in Section \ref{section:task} (see also Frank, 2014\footnote{\url{http://babieslearninglanguage.blogspot.com/2014/ 02/psychological-plausibility-considered.html}}).

\section{Can reverse engineering address the fully interactive learning scenario?}\label{section:interactive}

In this section, we revisit the simplifying assumptions of the input-driven scenario endorsed in Section \ref{section:feasibility} and displayed in Figure \ref{fig:relaf1}a. This scenario does not take into consideration the child's output, nor the possible feedback loops from the parents based on this output. Many researchers would see this as a major, if not fatal, limitation of the approach. In real learning situations, infants are also agents, and the environment reacts to their outputs creating feedback loops \cite{bruner_ontogenesis_1975,bruner_childs_1983,macwhinney_87,snow_mothers_1972,tamis2008parents}. 

\begin{figure}
	\centering
	\includegraphics[height=4.5cm]{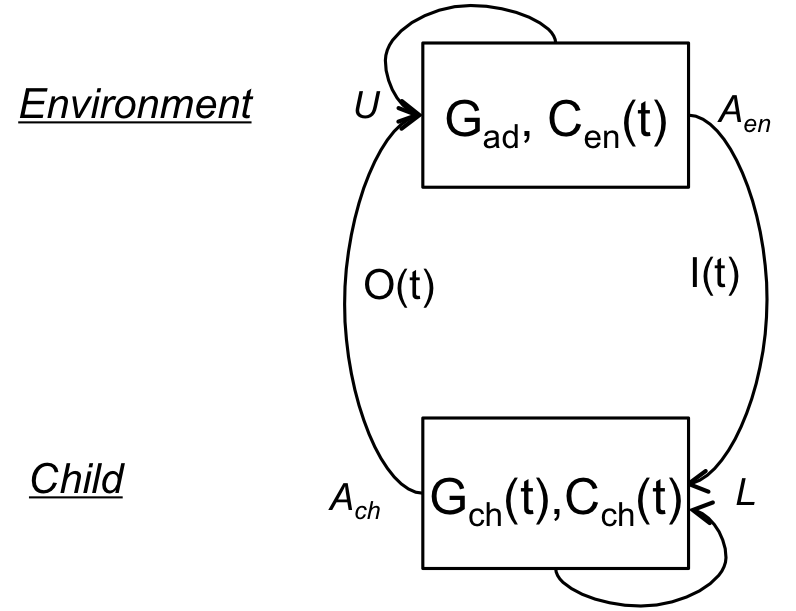}
	\caption{The learning situation in the interactive scenario, viewed as two coupled dynamic systems: the Child and the Environment.}
	\label{fig:relaf2}
\end{figure}

The most general description of the learning situation is therefore as in Figure \ref{fig:relaf2}. Here, the child is able to generate observable actions (some linguistic, some not) that will modify the internal state of the environment (through the monitoring function). The environment is able to generate the input to the child as a function of his internal state. In this most general form, the learning situation consists therefore in two \emph{coupled dynamic systems}.\footnote{We thank Thomas Schatz, personal communication, for proposing this general formulation.} 

Could such a complex situation be addressed within the reverse engineering approach? We would like to answer with a cautious yes, to the extent that it is possible to adhere to the same three requirements, i.e., realistic data (as opposed to simplified ones),  explicit criteria of success (based on cognitive indistinguishability), and scalable modeling (as opposed to verbal theories or toy models).   While none of these requirements seem out of reach, we would like to pinpoint some of the difficulties, which are the source of our caution. 

Regarding the data, the interactive scenario would require accessing the full (linguistic and non linguistic) output of the infant, not only her input. While this is not intrinsically harder to collect than the input, and is already been done in many corpora for older children, the issue of what to categorize as linguistic and non linguistic output and how to annotate it  is not completely trivial.  

Regarding computational modeling, instead of focusing on only one component (the learner) of one agent (the child), in the full interactive framework, one has to model two agents (the child and the adult) for a total of four components (the learner, the infant generator, the caregiver monitor, and the caregiver generator). Furthermore, the internal states of each agent has to be split into linguistic states (grammars) and non-linguistic (cognitive) states to represent the communicative aspects of the interaction (e.g., communicative intent, emotional/reinforcement signals). This, in turn, causes the split of each processing component into linguistic and cognitive subcomponents.

Although this is clearly a difficult endeavor, many of the individual ingredients needed for constructing such a system are already available in the following research areas. First, within speech technology, there are available components to build a language generator, as well as the perception and comprehension components in the adult caregiver. Second, within linguistics, psycholinguistics and neuroscience, there are interesting theoretical models of the learning of speech production and articulation in young children \cite{tomasello_03,johnson:2010,guenther_2012}. Third, within machine learning, great progress has been made recently on reinforcement learning, a powerful class of learning algorithms which assume that besides raw sensory data, the environment only provides sporadic positive or negative feedback \cite{sutton_1998}. This could be adapted to model the effect of the feedback loops on the learning components of the caregiver and the infant. Fourth, developmental robotics studies have developed the notion of intrinsic motivation, where the agent actively seek new information by being reinforced by its own learning rate \cite{oudeyer_2007}. This notion could be used to model the dynamics of learning in the child, and the adaptive effects of the caregiver-child feedback loops.

The most difficult part of this enterprise would perhaps concern the evaluation of the models. 
Indeed, each of these new components and subcomponents would have to be evaluated on their own in the same spirit as before, i.e., by running them on scalable data and testing them using human-validated tasks. For instance, the child language generator should be tested by comparing its output to age appropriate children's outputs, which requires the development of appropriate metrics (sentence length, complexity, etc) or human judgments. The cognitive subcomponents would have to be tested against experiments studying children and adults in experimentally controlled interactive loops \cite<e.g., >{smith_2008,goldstein_2008}. In addition, because a complex system is more than the sum of its parts, individual component validation would not sufficient, and the entire system would have to be evaluated.\footnote{For instance, a combined learner/caregiver system should be able to converge on a similar grammar as a learner ran on realistic data. In addition, their interactions should not differ in 'naturalness' compared to what can be recorded in natural situations, see \cite{bornstein_2010}.}

Fully specifying the methodological requirements for the reverse engineering of the interactive scenario would be a project of its own. It is not clear at present how much of the complications introduced by this scenario are necessary, at least to understand the first steps of language bootstrapping. To the extent that there are cultures where the direct input to the child is severely limited and/or the interactive character of that input circumscribed, it would seem that a fair amount of bootstrap can take place outside of interactive feedback loops. This is of course entirely an empirical issue, one that the reverse engineering approach should help to clarify.

\setcounter{secnumdepth}{0} %in order to not number bibliography
\bibliography{quantpaper}

\end{document}